\documentclass[10pt,journal,compsoc]{IEEEtran}
\ifCLASSOPTIONcompsoc
  \usepackage[nocompress]{cite}
\else
  \usepackage{cite}
\fi
\ifCLASSINFOpdf
\else
\fi
\usepackage{array}
\usepackage{amssymb}
\usepackage{amsmath}
\usepackage{booktabs}
\usepackage{graphicx}
\usepackage{xcolor}
\usepackage{amsthm}
\usepackage{multirow}
\usepackage{float}
\usepackage{hyperref}
\usepackage{blindtext}
\usepackage{tablefootnote}
\usepackage{nicefrac,xfrac}

\newtheorem{theorem}{Theorem}

\newcommand{\SQ}[1]{{\color{red}SQ: #1}}

\begin{document}
%
\title{Neural-IMLS: Self-supervised Implicit Moving Least-Squares Network for Surface Reconstruction}
%
%
%
%

\author{Zixiong Wang,
        Pengfei Wang,
        Pengshuai Wang,
        Qiujie Dong,
        Junjie Gao,
        Shuangmin Chen,
        Shiqing Xin,
        Changhe Tu,
        Wenping Wang~\IEEEmembership{Fellow,~IEEE}
\IEEEcompsocitemizethanks{\IEEEcompsocthanksitem 
Zixiong Wang, Pengfei Wang, Qiujie Dong, Junjie Gao, Shiqing Xin, and Changhe Tu, are with the School of
Computer Science and Technology, Shandong University. E-mail: zixiong\_wang@outlook.com, 8144756@qq.com, qiujie.jay.dong@gmail.com, gjjsdnu@163.com, xinshiqing@sdu.edu.cn, chtu@sdu.edu.cn.

\IEEEcompsocthanksitem Pengshuai Wang is with the Wangxuan Institute of Computer Technology, Peking University.
E-mail: wangps@hotmail.com.

\IEEEcompsocthanksitem Shuangmin Chen is with the School of Information and Technology, Qingdao University of Science and Technology. E-mail: csmqq@163.com.

\IEEEcompsocthanksitem Wenping Wang is with the Department of Computer Science and Engineering, Texas A\&M University. E-mail: wenping@tamu.edu.

}
\thanks{The corresponding author: xinshiqing@sdu.edu.cn \\ Manuscript received April 19, 2005; revised August 26, 2015.}}

%
%

\markboth{Journal of \LaTeX\ Class Files,~Vol.~14, No.~8, August~????}%
{Shell \MakeLowercase{\textit{et al.}}: Bare Demo of IEEEtran.cls for Computer Society Journals}
%



\IEEEtitleabstractindextext{%
\begin{abstract}
Surface reconstruction is very challenging when the input point clouds,
particularly real scans, are noisy and lack normals.
Observing that the Multilayer Perceptron (MLP) and the implicit moving least-square function (IMLS) provide a dual representation of the underlying surface, we introduce {\em Neural-IMLS}, 
a novel approach that directly learns the noise-resistant signed distance function (SDF) from unoriented raw point clouds in a self-supervised fashion. 
We use the IMLS to regularize the distance values reported by the MLP
while using the MLP to regularize the normals of the data points for running the IMLS. 
We also prove that at the convergence,
our neural network, benefiting from the mutual learning mechanism between the MLP and the IMLS, produces a faithful SDF whose zero-level set approximates the underlying surface. 
We conducted extensive experiments on various benchmarks, including synthetic scans and real scans.
The experimental results show that {\em Neural-IMLS} can reconstruct faithful shapes on various benchmarks with noise and missing parts. 
The source code can be found at~\url{https://github.com/bearprin/Neural-IMLS}.
\end{abstract}

}

\maketitle

\IEEEdisplaynontitleabstractindextext

%
\IEEEpeerreviewmaketitle

\IEEEraisesectionheading{\section{Introduction}\label{sec:introduction}}
\IEEEPARstart{S}{urface} reconstruction from an unstructured point cloud remains an active research topic since it is essential in many downstream computer vision and graphics applications, such as games, rendering, and animation.
The target surface is assumed to be 2-manifold and watertight in most scenarios.
Because of this, the signed distance function~(SDF)~$\boldsymbol{f}$ serves as a popular representation of the surface reconstruction problem. 
Given a point cloud~$\boldsymbol{P}$, the SDF~$\boldsymbol{f}$ to be reconstructed has to satisfy two basic conditions including
(1)~all the points are nearly sitting on the zero-value level-set surface, i.e., $\boldsymbol{f}(\boldsymbol{p}_i)\approx 0$ for each point~$\boldsymbol{p}_i$ in~$\boldsymbol{P}$, and (2)~the gradients of $\boldsymbol{f}$ are approximated as unit vector almost everywhere. 

\begin{figure}[!htp]
    \centering
    \includegraphics[width=0.95\linewidth]{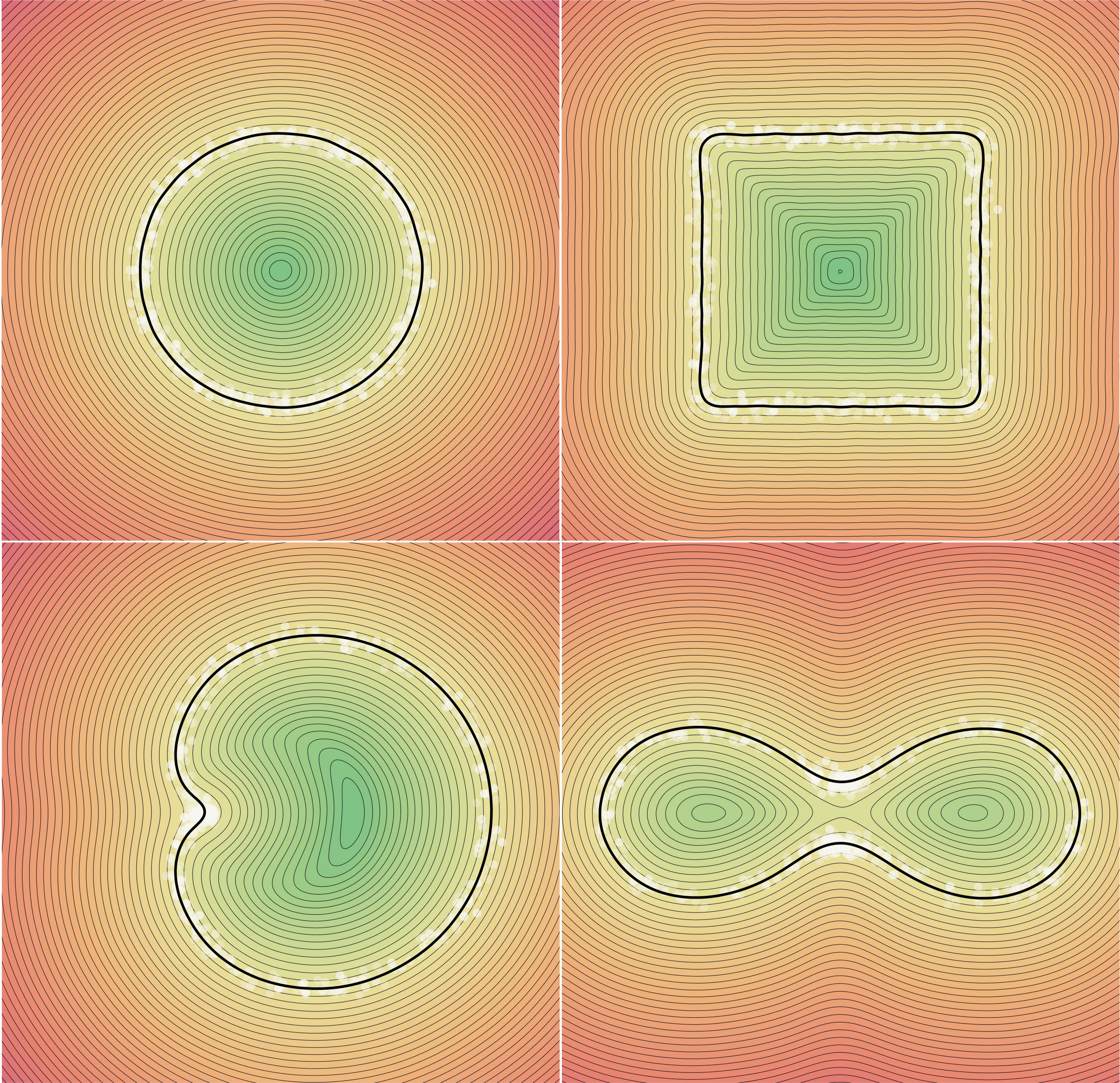}
    \caption{Taking the unoriented and noisy points  (the white points) as the input,
    {\em Neural-IMLS} learns a noise-resistant signed distance function~(SDF) whose zero level set~(the black lines) reports the reconstructed surface.
    }
    \label{fig:2d_level}
\end{figure}

Due to the fact that the raw data inevitably contains noise and sometimes lacks normals,
it is notoriously hard yet fascinating to reconstruct a high-fidelity surface in the presence of severe noise. 
The difficulties are two-fold.
On the one hand, one has to sacrifice the first condition to deal with noise while utilizing the positional clues of the noisy points as much as possible, which
necessitates a careful balance.
On the other hand, it is non-trivial to recover the real surface variations, especially when severe noise exists. 
The theme of this paper is to study the surface reconstruction problem assuming that 
the input point cloud is noisy and lacks normals, without any supervising data.

\begin{figure}[!htp]
    \centering
    \includegraphics[
      width=.45\textwidth
    ]{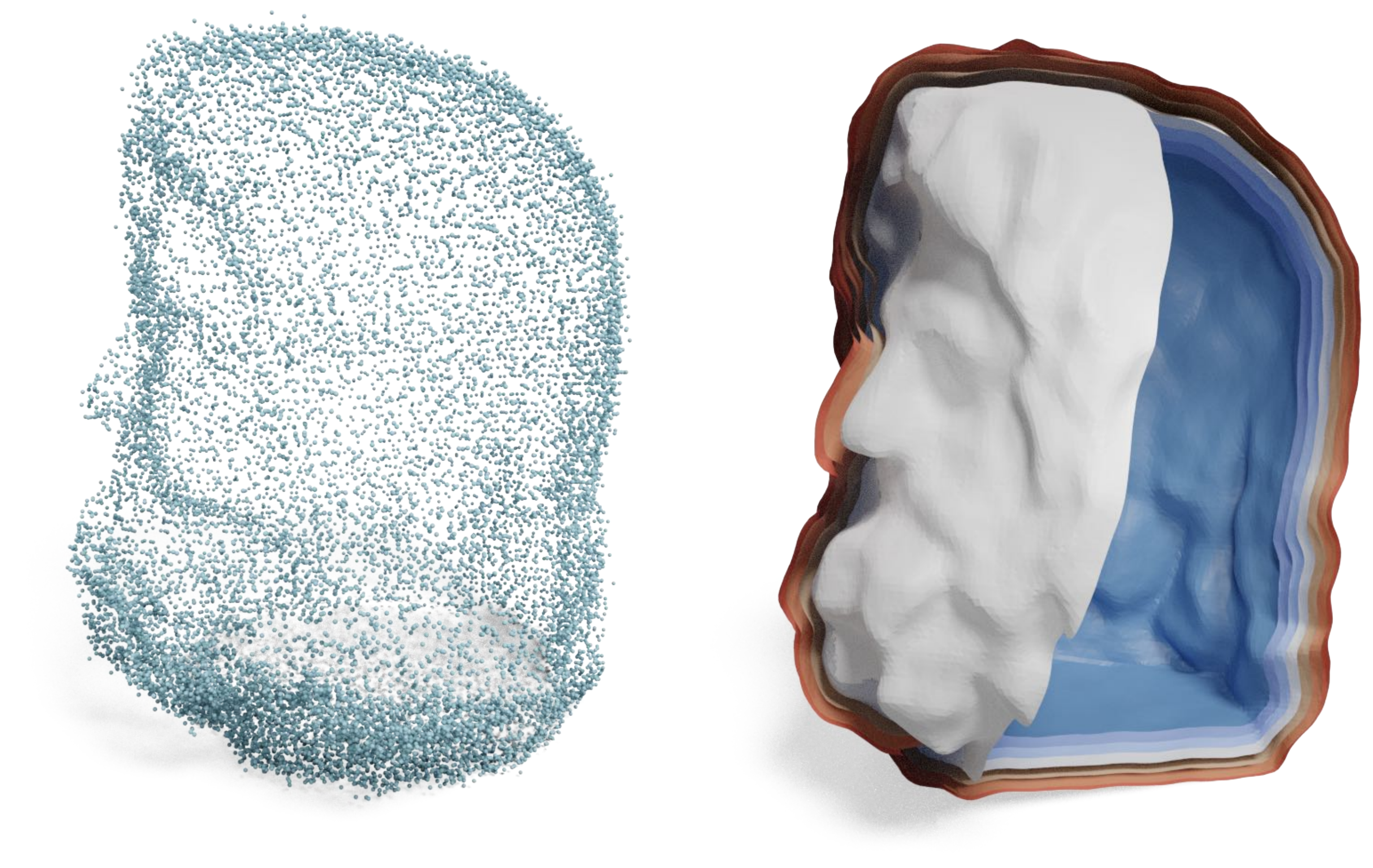}
    \vspace{-3mm}
    \caption{
    Our network can directly predict a high-fidelity surface (right) from a noisy input point cloud (left).
    We also visualize the inward and outward offset surfaces by extracting the iso-value level sets at different values. }
    \label{fig:level}
\end{figure}
Most traditional implicit approaches fit an implicit function by leveraging oriented normals~\cite{kolluri2008imls, oztireli2009RIMLS, kazhdan2006poisson, kazhdan2013screened}, followed by extracting the zero-level surface (e.g., Marching Cubes~\cite{lewiner2003efficient}). 
It's observed that the oriented normals have a decisive influence on inferring the variation of the final reconstructed surface. 
They cannot work except that the normals are given. 
To our knowledge, 
there are several traditional surface reconstruction approaches for handling unoriented point clouds, such as VIPSS~\cite{huang2019vipss} and iPSR~\cite{hou2022iterative},
but these approaches lack robust noise resistance.

Many supervised learning-based algorithms have been proposed for implicit surface reconstruction in recent years~\cite{xie2022neuralsurvey}.
Most of them learn an implicit function by fitting data samples supervised with the ground-truth SDFs or Occupancy Fields~\cite{park2019deepsdf, mescheder2019onet, peng2020convolutional, jiang2020local, erler2020points2surf, Liu2021MLS, Boulch2022poco}. 
However, these methods can not be used when the ground-truth SDFs or Occupancy Fields are hard to compute. 
There are some deep learning-based methods~\cite{IGR, SAL, sitzmann2020siren, atzmon2021sald, NeuralPull, ben2022digs, ma2022surface} 
that do not require knowledge of ground-truth field values.
When severe noise exists, however, they lack a smart mechanism to tune the degree to which the input points should be enforced on the zero-level surface.

We have made a couple of interesting observations.
On the one hand, the implicit moving least-square function (IMLS)~\cite{kolluri2008imls,oztireli2009RIMLS} can accurately reconstruct the underlying surface by estimating SDFs near the surface, only if the point normals are provided.
On the other hand, MLPs fit the input cloud globally to reconstruct the SDF and can robustly provide the normal vector field by automatic differentiation, 
whereas it is also known that MLPs are hard to recover fine-scale geometric features and often produce over-smoothed surfaces.

\begin{figure*}[!htp]
    \centering
    \includegraphics[width=.8\textwidth]{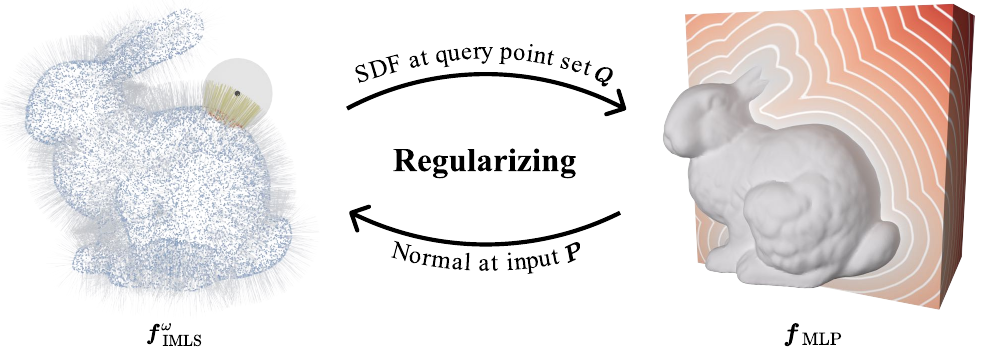}
    \caption{
    Considering that MLP and IMLS as a dual representation 
    of the underlying SDF, we train the neural network by mutual regularization between the MLP $\boldsymbol{f}_{\text{MLP}}$ and the IMLS $\boldsymbol{f}_{\text{IMLS}}^{\omega}$:
    MLP estimates point normals for IMLS by regressing SDFs globally, and IMLS estimates accurate SDFs near the surface and help MLPs to recover shape features near the surface.
    }
    \label{fig:insight}
\end{figure*}

Based on the observations, we propose {\em Neural-IMLS}, a novel method for learning the SDF directly from poor-quality (noisy, irregular, and without normals) point clouds (see Fig.~\ref{fig:2d_level},~\ref{fig:level}).
Our method learns the underlying SDF in a self-supervised fashion by minimizing the loss between a couple of SDFs, one obtained by the implicit moving least-square function (IMLS) and the other by the Multilayer Perceptron (MLP).
The MLP and the IMLS are taken as a dual representation of the underlying surface since they complement each other: 
the IMLS regularizes SDFs near the surface and helps the MLP to reconstruct geometric details and sharp features by providing the estimated SDFs as the regression target of the MLP, 
while the MLP regularizes and estimates the normals of the points and enables the running of the IMLS by providing the estimated normals to the IMLS.
In this way, the two SDFs, respectively given by the MLP and the IMLS, promote each other during neural network optimization.
We also define a novel loss function by considering the difference between the two SDFs
and the spatial coherence of the gradients at the same time.
The rationale can be understood based on the mechanism of mutual learning~\cite{gou2021knowledge} (see Fig.~\ref{fig:insight}).
After optimization convergence, the neural network is able to learn the SDF that encodes the underlying shape accurately. 

%
%

We conducted extensive experiments on a large variety of shapes, including synthetic scans and real scans. 
The experimental results (see Section~\ref{sec:exp}) demonstrate that our approach can learn more accurate SDFs than other leading overfitting techniques, especially on noisy scans. 
The main contributions of this paper are:
\begin{itemize}
\item[$\bullet$] A self-supervised neural network for reconstructing a 2-manifold surface from
a noisy and unoriented input point cloud. The key insight is that the MLP and the IMLS can be taken as a dual representation of the underlying surface. Our network is provably convergent. 
 \item[$\bullet$] A novel loss that considers not only the spatial coherence of the gradients but also the difference between a pair of distance fields, one reported by the IMLS and the other reported by the neural network.
 \item[$\bullet$] Extensive comparative experiments and ablation studies to validate the effectiveness of our approach on various benchmarks, including both synthetic and real-scan data.
\end{itemize}

\section{Related Work}
Surface reconstruction is a fundamental research topic in computer graphics and computer vision.
Numerous reconstruction algorithms~\cite{huang2022survey} have been proposed in the last three decades. 
In this section, we briefly review the implicit surface reconstruction methods, including traditional and learning-based methods.

\subsection{Traditional Implicit Methods} 
Most traditional implicit methods require the input points to be equipped with orienting normals. 
Some methods~\cite{carr2001reconstruction, li2016sparse, huang2019vipss} leverage the radial basis function (RBF) to represent the underlying signed distance function (SDF) as a weighted combination of a group of radial basis kernels.
Poisson reconstruction~\cite{kazhdan2006poisson} and its
variants~\cite{kazhdan2013screened, kazhdan2020poisson} seek a continuous occupancy field that can be formulated as the solution to Poisson's equation. 
Besides, the implicit moving least-square methods (IMLS)~\cite{shen2004interpolating, kolluri2008imls, oztireli2009RIMLS, schroers2014HessianIMLS}
approximates the underlying SDF by blending local smooth planes.

Generally speaking, the traditional implicit reconstruction approaches need to take orienting normals as the input, and thus the reconstruction quality depends heavily on the reliability of normal vectors.
However, estimating orienting normal vectors is highly non-trivial, especially for point clouds with severe defects (e.g., noise, missing regions, and thin plates). 
It's worth noting that although there are a few traditional algorithms~\cite{hou2022iterative} that can work without normals, they lack the capacity for noise resistance. 

\subsection{Learning Implicit Function with Ground Truth Supervision}
Recently, several learning-based methods have been proposed to learn implicit functions from a large dataset with ground-truth SDFs or occupancy values~\cite{xie2022neuralsurvey}. 
Early works tend to encode each shape into a fixed-length latent code and then recover the underlying surface by a decoding
operation~\cite{park2019deepsdf, mescheder2019onet}.
While these methods can encode the overall shape for a group of similar models, they cannot deal with unseen shapes whose geometric and topological structures differ greatly from the training data. 
Therefore, some local feature-based methods have been proposed to address this issue.
LIG~\cite{jiang2020local} splits the input point cloud into point patches 
and learns the patch-wise geometric features across various shapes. 
Peng et al.~\cite{peng2020convolutional} proposed a convolutional operator for aggregating local and global information. 
In the decoding phase, this method leverages grid features with 3D-UNet to interpolate features for query points.
Points2surf~\cite{erler2020points2surf} uses global features to predict signs, and local features to predict distances, respectively. 
It leverages a mechanism called sign propagation to reduce the computational cost of the inference phase. 
However, the sign propagation yields inaccurate results when the input point cloud does not have a high-quality distribution (as pointed out in~\cite{yifan2020isopoints}), causing unwanted holes in the reconstructed mesh.
By combining the benefits of the implicit approaches with the point set methods, DeepMLS~\cite{Liu2021MLS} can generate a set of MLS projection points that help predict the implicit approximation surface. 
Unfortunately, DeepMLS cannot learn reliable parameters (e.g., radius) that facilitate the computation of IMLS, especially for sparse and noisy point clouds.
It is likely to produce unwanted ghost geometries.
POCO~\cite{Boulch2022poco} proposes a kind of attention mechanism convolution to compute latent vectors at each input point, which is scalable to scenes of arbitrary size.
Recently, some methods~\cite{tang2021octfield, wang2022dual} 
are proposed to partition and encode the given point cloud by an octree,
and then decode the field from the feature code of octree nodes.
The methods mentioned above, whether global or local, need supervision with the ground-truth SDF or occupancy function. 
The supervised learning methods~\cite{neyshabur2017exploring}, however,
generally need a huge amount of training data as the input and a considerable timing cost to converge, and may not generalize well on the shapes that are not available in the training set.


\begin{figure*}[!htp]
    \centering
    \includegraphics[
      width=\textwidth,
    ]{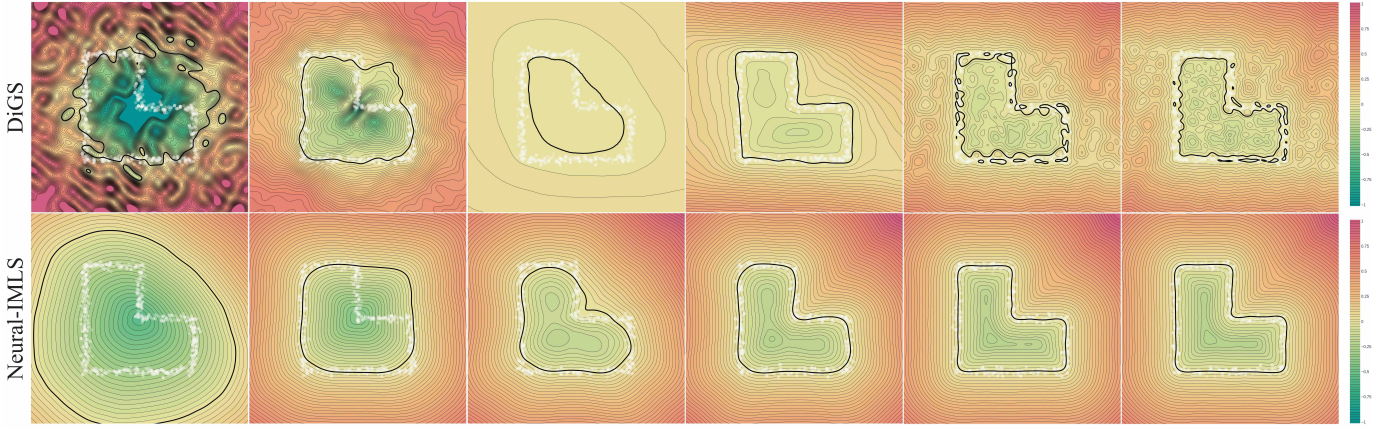}
    \vspace{-3mm}
    \caption{
    Comparison between DiGS~\cite{ben2022digs} (top) and ours (bottom)
    on how the learned SDF changes with the increasing number of iterations for the noisy input shape L.
    }
    \label{fig:2diter}
\end{figure*}

\subsection{Learning Implicit Function from Raw Data}
There are some approaches aiming at learning an implicit function from raw point clouds without knowledge of ground truth SDF or occupancy. 
They generally introduce some principles to constrain the predicted implicit function (e.g., gradients being a unit vector). 
SAL/SALD~\cite{SAL, atzmon2021sald} performs sign agnostic regression to get a signed version from the unsigned distance function. 
IGR~\cite{IGR} introduces a geometric regularization term to encourage the predicted implicit function to have unit gradients, which can work with or without normals.
Ma et al.~\cite{NeuralPull} trained a neural network to predict the signed distance as well as the gradients so that a query point can be pulled onto the underlying surface. 
Next, they also proposed PredictableContextPrior~\cite{ma2022surface} to combine local context prior and predictive context prior by an additional fine-tuning mechanism.
SAP~\cite{peng2021shape} introduces a differentiable 
a version of Poisson Surface Reconstruction to
repeatedly minimize the Chamfer distance between an explicit mesh and the input point cloud.
Lipman~\cite{lipman2021phase}
observed that there is a connection between the occupancy field and the SDF,
and thus introduced a loss to learn a density function whose log transform approaches the SDF, which is based on the theory of phase transitions.
DiGS~\cite{ben2022digs} considers a kind of soft constraint about the divergence of the gradients of the distance field in the loss function and can help produce more reliable results in situations where normals are not available. 
To summarize, most of them cannot produce high-quality reconstructed results 
in the presence of severe noise,
since it is hard to strike a careful balance
between respecting the positional clues and eliminating noise.

\section{Method}
First, we will state problems and review the definition of the IMLS surface in Section~\ref{ssec:problem} and Section~\ref{ssec:imls}, respectively. 
We then elaborate on the mutual-regularization mechanism of our method in Section~\ref{ssec:neural}.

\subsection{Problem Statement and Motivation}\label{ssec:problem}
Given an input point cloud
$\boldsymbol{P} = \left\{\boldsymbol{p}_i\right\}_{i \in I} \subset \mathbb{R}^{3}$, 
we aim to learn a signed distance function $\boldsymbol{f}$: $\mathbb{R}^{3} \rightarrow \mathbb{R}$, from which we can reconstruct a 2-manifold surface $\boldsymbol{S}$
by extracting the zero-value level set, i.e.,
\begin{equation}
    \boldsymbol{S} = \left\{\boldsymbol{q}\ \big | \ \boldsymbol{f}(\boldsymbol{q}) = 0, \boldsymbol{q} \in \mathbb{R}^3 \right\}.
\end{equation}




We consider the reconstruction problem from a different perspective.
We assume that $\mathcal{R}$ be a ``perfect'' surface reconstruction solver.
By ``perfect'', we mean that $\mathcal{R}$ takes the points $\boldsymbol{P} = \left\{\boldsymbol{p}_i\right\}_{i \in I} $ and the normals 
$\boldsymbol{N} = \left\{\boldsymbol{n}_i\right\}_{i \in I} $ as the input and is able to reconstruct the ``ideal'' surface $\boldsymbol{S}$
from the pair~$(\boldsymbol{N}, \boldsymbol{P})$. 
Let $\boldsymbol{N}^{(0)}$ be the initial orienting normals.
We consider the following iterative process:
\begin{equation}
     \begin{split}
    (\boldsymbol{N}^{(0)};~\boldsymbol{P}) \stackrel{\mathcal{R}}\longrightarrow \boldsymbol{S}^{(0)}\longrightarrow (\boldsymbol{N}^{(1)};~\boldsymbol{P})\stackrel{\mathcal{R}}\longrightarrow \boldsymbol{S}^{(1)} \\ \longrightarrow,\cdots,
    \stackrel{}\longrightarrow (\boldsymbol{N}^{(\infty)};~\boldsymbol{P})\stackrel{\mathcal{R}}\longrightarrow \boldsymbol{S}^{(\infty)},
    \label{eq::obs}
     \end{split} 
\end{equation}
where $\boldsymbol{S}^{(i)}$ 
denotes the underlying surface
determined by applying the ``perfect'' reconstruction solver $\mathcal{R}$ 
on the pair $(\boldsymbol{N}^{(i)}, \boldsymbol{P})$,
and $\boldsymbol{S}^{(i)}$ can help update the setting of normals in the iteration.
Obviously, when the initial normals $\boldsymbol{N}^{(0)}$ coincidentally align with the ``ideal'' target surface, the surface $\boldsymbol{S}^{(0)}$ is exactly the same with the target surface,
and thus repeating the iterations cannot produce any change to~$\boldsymbol{S}^{(0)}$.
In real scenarios, however,
the ``perfect'' reconstruction solver,
as well as the ``ideal'' normals, are not available.
We suggest using the IMLS 
to drive the evolution of the implicit representation of~$\boldsymbol{S}^{(i)}$.
We hope that both the normals and the underlying surface
can progressively change toward the ``ideal''  configuration
during mutual update between the IMLS and the neural network.

\begin{figure*}[!htp]
    \centering
    \includegraphics[
      width=0.8\textwidth
    ]{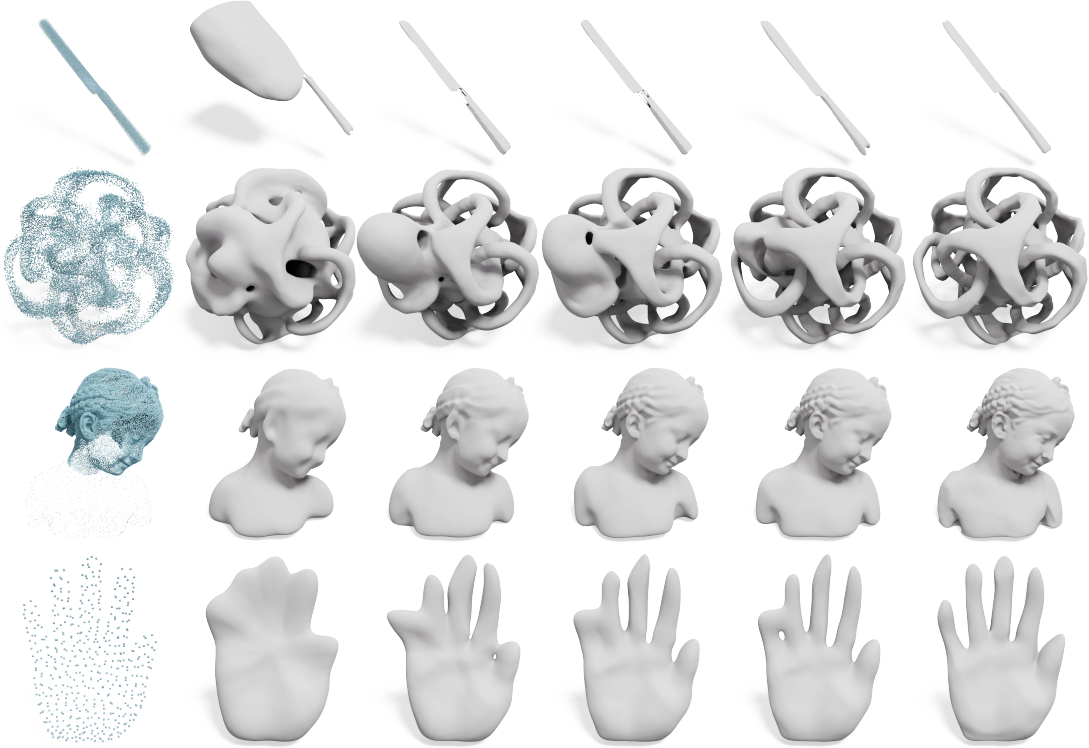}
    \vspace{-3mm}
    \caption{
    We visualize the iterative process for four models with various artifacts at the 1st,
    10th, 30th, 50th, and 100th iterations, respectively.
    From top to bottom: a misaligned point cloud of the thin-structure Knife model,
    a noisy point cloud of the high-genus model,
    {a non-uniform point cloud of the Bimba model with rich geometric details},
    and a super-sparse point cloud of the Hand model with thin gaps between fingers. 
    In spite of various artifacts, our method can still reconstruct a high-fidelity surface.
    \label{fig:3diter}}
\end{figure*}

\subsection{IMLS Surface}\label{ssec:imls}
Each point $\boldsymbol{p}_i\in \boldsymbol{P}$
contributes to the query point~$\boldsymbol{q}$
a signed distance $\left\langle\boldsymbol{q}-\boldsymbol{p}_i, \boldsymbol{n}_i\right\rangle$,
where $\boldsymbol{n}_i$ is the unit normal vector at~$\boldsymbol{p}_i$.
The final distance at~$\boldsymbol{q}$
is defined as a weighted average~\cite{kolluri2008imls}:
\begin{equation}
    \boldsymbol{f}^{\boldsymbol{P}}_\text{IMLS}(\boldsymbol{q};\boldsymbol{P},\boldsymbol{N}) =\frac{\sum_{\boldsymbol{p}_i \in \boldsymbol{P}} \theta\left(\left\|\boldsymbol{q}-\boldsymbol{p}_i\right\|\right) \cdot\left\langle\boldsymbol{q}-\boldsymbol{p}_i, \boldsymbol{n}_i\right\rangle}{\sum_{\boldsymbol{p}_i \in \boldsymbol{P}} \theta\left(\left\|\boldsymbol{q}-\boldsymbol{p}_i\right\|\right)},
    \label{eq::imls_ori}
\end{equation}
where $\theta\left(\left\|\boldsymbol{q}-\boldsymbol{p}_i\right\|\right)$ denotes the weighting scheme and generally gives the nearer points in~$\boldsymbol{P}$ a bigger influence. 
By default, $\theta$ is set as the Gaussian function $\theta(x)=\exp({-x^{2} /\sigma_\text{IMLS}^{2}})$, where $\sigma_\text{IMLS}$ is the support radius and theoretically depends on the local feature size~(LFS).
The IMLS has various forms, among which the IMLS scheme is the simplest case of~\cite{shen2004interpolating}.



Since we hope that the zero-value level set of~$\boldsymbol{f}^{\boldsymbol{P}}_\text{IMLS}$ defines the target surface,
$\boldsymbol{f}^{\boldsymbol{P}}_\text{IMLS}$ should report an as-accurate-as-possible
distance value for points nearby the underlying surface.
To be more specific, we constrain the query point~$\boldsymbol{q}$ to be lying in the thin shell
of a width~$2r$. 
Let $\boldsymbol{B}_r(\boldsymbol{q})$ be the open ball centered at~$\boldsymbol{q}$. 
By ignoring the contribution of those points outside~$\boldsymbol{B}_r(\boldsymbol{q})$
(note that $\theta$ decreases with the increasing distance between~$\boldsymbol{q}$ and $\boldsymbol{p}_i$),
Eq.~(\ref{eq::imls_ori}) can be reduced to:
\begin{equation}
\begin{aligned}
    &\boldsymbol{f}^r_\text{IMLS}(\boldsymbol{q};\boldsymbol{B}_{r}(\boldsymbol{q}),\boldsymbol{N}) \\ 
    &=\frac{\sum_{\boldsymbol{p}_i \in \boldsymbol{B}_{r}(\boldsymbol{q})} \theta\left(\left\|\boldsymbol{q}-\boldsymbol{p}_i\right\|\right) \cdot\left\langle\boldsymbol{q}-\boldsymbol{p}_i, \boldsymbol{n}_i\right\rangle}{\sum_{\boldsymbol{p}_i \in \boldsymbol{B}_r(\boldsymbol{q})} \theta\left(\left\|\boldsymbol{q}-\boldsymbol{p}_i\right\|\right)}.
    \label{eq::imls_ori_local}
\end{aligned}
\end{equation}
Eq.~(\ref{eq::imls_ori_local}) has two aspects of benefits.
First, it requires less computational overhead than Eq.~(\ref{eq::imls_ori}).
Second, it encourages~$\boldsymbol{f}^r_\text{IMLS}$ to be estimated based on local information.

\subsection{Insight}\label{ssec:Insight}
As pointed out in~\cite{kolluri2008imls},
the IMLS has a noise-resistance ability.
It can predict a noise-resistant distance field
if the normals are available. 
Inspired by this, we use the neural network~$\boldsymbol{f}^r_\text{MLP}$ to encode the implicit surface,
allowing the IMLS to help tune $\boldsymbol{f}^r_\text{MLP}$
such that the SDF and gradients of $\boldsymbol{f}^r_\text{MLP}$ are compatible.
During the regularization, $\boldsymbol{f}^r_\text{MLP}$ can 
learn some knowledge about the distance values given by the IMLS,
and the IMLS can also be promoted by the gradients of~$\boldsymbol{f}^r_\text{MLP}$,
which is the following iterative process:
\begin{equation}
     \begin{split}
    (\boldsymbol{N}^{(0)};~\boldsymbol{P}) \stackrel{\text{IMLS}}\longrightarrow \boldsymbol{f}_\text{MLP}^{(0)}\longrightarrow (\boldsymbol{N}^{(1)};~\boldsymbol{P})\stackrel{\text{IMLS}}\longrightarrow \boldsymbol{f}_\text{MLP}^{(1)} \\ \longrightarrow,\cdots,
    \stackrel{}\longrightarrow (\boldsymbol{N}^{(\infty)};~\boldsymbol{P})\stackrel{\text{IMLS}}\longrightarrow \boldsymbol{f}_\text{MLP}^{(\infty)}.
    \label{eq::insight}
     \end{split} 
\end{equation}
We define a smart mechanism for tuning the neural network~$\boldsymbol{f}_\text{MLP}$ to the ``ideal'' configuration using the IMLS.

\vspace{.5mm}
\noindent\textbf{Remark.}
It can be imagined that one can use the IMLS purely to define a similar iterative process, i.e., using the gradients of the IMLS distance field to update the normals of the point cloud. 
Recall that the theme of this paper is to study the surface reconstruction problem assuming that the input point cloud is noisy and lacks normals, and contains a limited number of points.
For this purpose, we utilize the MLP to encode the spatial relations inherent in the point cloud and robustly estimate normals for IMLS with the MLP.
Furthermore, although a pure MLP can also be trained on a noisy point cloud without normals, the reconstructed surfaces are often over-smoothed. 
The SDFs estimated by IMLS can help MLP to improve its fitting accuracy near the surface, thus producing better results than a pure MLP.
We conduct an ablation study about the benefits of our method over a pure IMLS or MLP scheme in Section~\ref{ssec:ablation}.

\subsection{Neural-IMLS}\label{ssec:neural}
We propose to use the IMLS to regularize the distance values reported by the MLP
while using the MLP to regularize the normals of the data points for running the IMLS. 
We hope that the neural network can output faithful SDF whose zero-level set well approximates the underlying surface, benefiting from the mutual learning mechanism.

\vspace{.5mm}
\noindent\textbf{Na\"{i}ve loss.}
Let $\boldsymbol{Q}$ be a set of pre-computed query points around $\boldsymbol{P}$,
$\boldsymbol{f}_\text{MLP}$ and $\boldsymbol{f}^r_\text{IMLS}$ be the two SDFs obtained by the MLP and the IMLS.
To promote the IMLS by the learned gradients from the neural network,
we need to feed the gradients of~$\boldsymbol{f}_\text{MLP}$
into the IMLS iterative scheme:
\begin{gather}
 \boldsymbol{f}^r_\text{IMLS}(\boldsymbol{q};\boldsymbol{B}_{r}(\boldsymbol{q}),\nabla \boldsymbol{f}_\text{MLP})\qquad\qquad\qquad\qquad\nonumber\\
    =\frac{\sum_{\boldsymbol{p}_i \in \boldsymbol{B}_{r}(\boldsymbol{q})} \theta\left(\left\|\boldsymbol{q}-\boldsymbol{p}_i\right\|\right) \cdot\left\langle\boldsymbol{q}-\boldsymbol{p}_i, \frac{\nabla \boldsymbol{f}_\text{MLP}(\boldsymbol{p}_i)}{\left\| \nabla \boldsymbol{f}_\text{MLP}(\boldsymbol{p}_i) \right\|} \right\rangle}{\sum_{\boldsymbol{p}_i \in \boldsymbol{B}_{r}(\boldsymbol{q})} \theta\left(\left\|\boldsymbol{q}-\boldsymbol{p}\right\|\right)},\label{eq::imls::learned}
\end{gather}
where $\nabla \boldsymbol{f}_\text{MLP}$ is computed based on the back-propagation of the neural network.
In fact, $\frac{\nabla \boldsymbol{f}_\text{MLP}(\boldsymbol{p}_i)}{\left\| \nabla \boldsymbol{f}_\text{MLP}(\boldsymbol{p}_i) \right\|}$ can be taken as the new
normal vector at the point~$\boldsymbol{p}_i$.

We thus define a loss by measuring the
squared difference between~$\boldsymbol{f}_\text{MLP}$ and $\boldsymbol{f}^r_\text{IMLS}$ as follows:
\begin{equation}
\label{eq::naive_loss:1}
L = \frac{1}{\left| \boldsymbol{Q} \right|} \sum_{\boldsymbol{q}_j \in \boldsymbol{Q}} \left\|\boldsymbol{f}^r_\text{IMLS}(\boldsymbol{q}_j)-\boldsymbol{f}_\text{MLP}(\boldsymbol{q}_j)\right\|_2 ^ 2,
\end{equation}
where ${\left| \boldsymbol{Q} \right|}$ is the number of query points in~$\boldsymbol{Q}$.

\vspace{.5mm}
\noindent\textbf{Spatial coherence loss.}
However, we observe that the reconstruction error near sharp features or thin geometric regions is relative high with this loss term.  
The reason is that the IMLS defined in Eq.~\ref{eq::imls::learned} only considers the spatial similarity of neighboring points when doing the weighted average.
To combat this issue, we introduce an influence function to measure the difference between
$\nabla\boldsymbol{f}_\text{MLP}(\boldsymbol{p})$ and $\nabla\boldsymbol{f}_\text{MLP}(\boldsymbol{q})$
for a pair of close points $\boldsymbol{p}$ and $\boldsymbol{q}$.

\begin{gather}
    \psi\left(\nabla \boldsymbol{f}_\text{MLP}(\boldsymbol{q}), \nabla \boldsymbol{f}_\text{MLP}(\boldsymbol{p})\right)\qquad\qquad\qquad\nonumber\\
    =\exp{\left(-\frac{\left\| \frac{\nabla \boldsymbol{f}_\text{MLP}(\boldsymbol{q})}{\left\| \nabla \boldsymbol{f}_\text{MLP}(\boldsymbol{q}) \right\|} - \frac{\nabla \boldsymbol{f}_\text{MLP}(\boldsymbol{p})}{\left\| \nabla \boldsymbol{f}_\text{MLP}(\boldsymbol{p}) \right\|}\right\|^2}{\sigma_\text{coherence}^2}\right)},\label{eq::gradients::similarity}
\end{gather}
where the hyperparameter $\sigma_\text{coherence}$ is used 
to tune how the spatial coherence depends on the similarity between gradients. 
We combine the two weightings schemes~$\theta$ and~$\psi$, 
and obtain a novel weighting scheme:
\begin{equation}
        \omega(\boldsymbol{q}, \boldsymbol{p}) = \theta\left(\left\|\boldsymbol{q}-\boldsymbol{p}\right\|\right)\psi(\nabla \boldsymbol{f}_\text{MLP}(\boldsymbol{q}),\nabla \boldsymbol{f}_\text{MLP}(\boldsymbol{p})). \\ 
\end{equation}
By plugging the combined weighting scheme into Eq.~(\ref{eq::imls::learned}), we have an improved version of IMLS:
\begin{gather}
    \boldsymbol{f}^\omega_\text{IMLS}(\boldsymbol{q};\boldsymbol{B}_{r}(\boldsymbol{q}),\nabla \boldsymbol{f}_\text{MLP})\qquad\qquad\qquad\qquad\nonumber\\
    =\frac{\sum_{\boldsymbol{p}_i \in \boldsymbol{B}_{r}(\boldsymbol{q})} \omega\left(\boldsymbol{q}, \boldsymbol{p}_i \right) \cdot\left\langle\boldsymbol{q}-\boldsymbol{p}_i, \frac{\nabla \boldsymbol{f_\text{MLP}(p_i)}}{\left\| \nabla \boldsymbol{f_\text{MLP}(p_i)} \right\|} \right\rangle}{\sum_{\boldsymbol{p}_i \in \boldsymbol{B}_{r}(\boldsymbol{q})} \omega\left(\boldsymbol{q}, \boldsymbol{p}_i \right)}.
    \label{eq:new_imls}
\end{gather}
Like Eq.~(\ref{eq::naive_loss:1}),
we re-define a loss by measuring the
squared difference between~$\boldsymbol{f}_\text{MLP}$ and $\boldsymbol{f}^\omega_\text{IMLS}$:
\begin{equation}
\label{eq::loss}
L = \frac{1}{\left| \boldsymbol{Q} \right|} \sum_{\boldsymbol{q}_j \in \boldsymbol{Q}} \left\|\boldsymbol{f}^\omega_\text{IMLS}(\boldsymbol{q}_j)-\boldsymbol{f}_\text{MLP}(\boldsymbol{q}_j)\right\|_2 ^ 2.
\end{equation}
By abuse of notation, in the following sections, we also use $\boldsymbol{f}_\text{IMLS}$ 
to represent the $\omega$-weighted IMLS iterative scheme.

\vspace{.5mm}
\noindent\textbf{Network architecture and training details.} 
We use the same architecture as IGR~\cite{IGR}: an MLP with 8 layers with 512 neurons in each layer (total 1.8M parameters), and take a single skip connection concatenating an input point as the input of the $4$-th layer.
We use the SoftPlus Activation Function ($\beta=1000$) and the geometric initialization (GNI) proposed by~\cite{SAL}, with which the network is initialized to approximate the SDF of the unit sphere. 
In each training iteration, we first sample a batch of query points $\boldsymbol{q}_j \in \boldsymbol{Q}$.
For each query point, we further sample its neighboring points in $\boldsymbol{B}_r(q_j)$.
Then we forward the MLP for these points to get the predicted gradients with automatic differentiation and evaluate $\boldsymbol{f}^\omega_\text{IMLS}(\boldsymbol{q}_j)$ according to Eq.~\ref{eq:new_imls} and then update the parameters of the MLP with the Adam optimizer according to the loss function defined in Eq.~\ref{eq::loss}.
When optimizing the MLP, $\boldsymbol{f}^\omega_\text{IMLS}(\boldsymbol{q}_j)$ in Eq.~\ref{eq::loss} is detached from the gradient computation.

To help understand the iterative process,
we visualize a 2D example and a 3D example in Fig.~\ref{fig:2diter} and Fig.~\ref{fig:3diter}, respectively. 
See Appendix~A for more details about training and sampling strategy.

\subsection{Proof}
\label{ssec:proof}
Next, we come to discuss why the learning function $\boldsymbol{f}_\text{MLP}$ can converge to a signed distance function. 

\begin{table}[!htp]
\centering
\caption{\textbf{Surface Reconstruction Benchmark.} We report Chamfer Distance (CD), Hausdorff distance (HD), and their mean deviation relative to the best-performing method (rel. CD and rel. HD). All methods do not require ground-truth supervision, and we evaluate them without normals.}
\label{table:srb}
\begin{tabular}{l|>{\centering}p{0.045\textwidth}>{\centering}p{0.05\textwidth}cc} 
\toprule
Method        & CD$\downarrow$         & HD$\downarrow$         & rel. CD         & rel. HD         \\ 
\midrule
SAL~\cite{SAL}          & 0.36          & 7.47          & 0.18          & 4.32           \\
IGR~\cite{IGR}          & 1.38          & 16.33         & 1.20          & 13.18           \\
IGR+FF~\cite{IGR}       & 0.96          & 11.06         & 0.78          & 6.33           \\
SIREN~\cite{sitzmann2020siren}        & 0.42          & 7.67          & 0.78          & 4.52          \\
PHASE+FF~\cite{lipman2021phase}     & 0.22          & 4.96          & 0.04          & 1.81           \\
SAP~\cite{peng2021shape}          & 0.20          & 4.60          & 0.02          & 1.45           \\
iPSR~\cite{hou2022iterative}         & 0.21          & 5.01          & 0.03          & 1.86           \\
DiGS~\cite{ben2022digs}         & 0.19          & 3.52          & 0.01          & 0.37           \\
PCP~\cite{ma2022surface}          & 0.54          & 7.11          & 0.36          & 3.96           \\
\textbf{Ours} & \textbf{0.18} & \textbf{3.15} & \textbf{0.00} & \textbf{0.00}  \\
\bottomrule
\end{tabular}
\end{table}
\begin{figure}[!htp]
    \centering
    \includegraphics[
      width=0.48\textwidth,
    ]{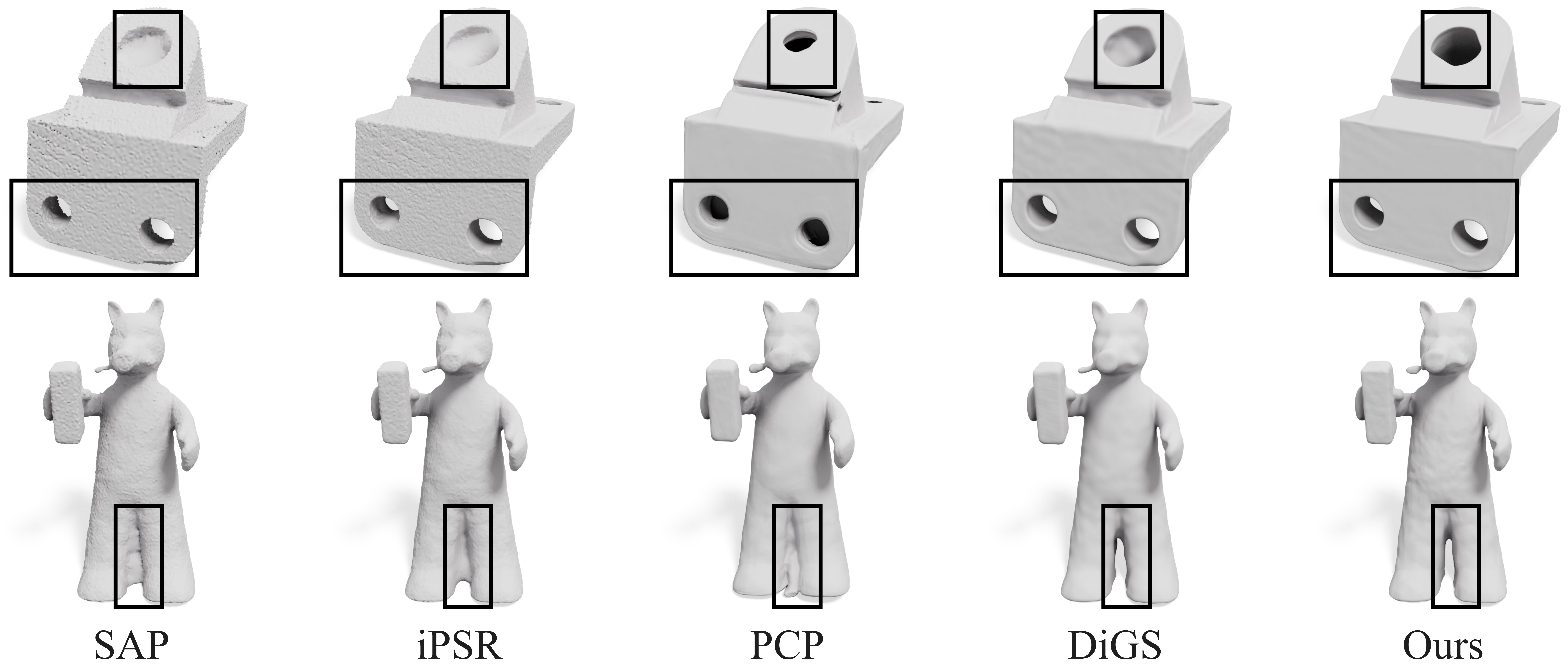}
    \vspace{-3mm}
    \caption{
    Qualitative comparison between different approaches on the Surface Reconstruction Benchmark~\cite{williams2019dgp}. These methods can work without normals and do not need supervision.
    It can be seen that our approach is comparable to DiGS and outperforms the other approaches on this benchmark.
    But we must point out that the five point clouds are not sparse, and the noise is not severe. DiGS has a good performance in this situation. }
    \label{fig:srb}
\end{figure}

\begin{theorem}
Suppose that the sample points are sufficiently dense, 
(e.g., satisfying the $\epsilon$-sampling condition~\cite{kolluri2008imls}). 
our network $\boldsymbol{f}_\text{MLP}$ converges to a \textit{signed} distance function near the surface.
\end{theorem}

\begin{proof}
For $\boldsymbol{f}^\omega_\text{IMLS}$ defined in Eq.~(\ref{eq:new_imls}), when a query point~$\boldsymbol{q}$ is sufficiently close to an input point $\boldsymbol{p}_i$ and $r\rightarrow 0$, the weight $\omega(\boldsymbol{q}, \boldsymbol{p_k})$ approaches 1 if~$\boldsymbol{p_k}=\boldsymbol{p_i}$ (or $k = i$), and $\omega(\boldsymbol{q}, \boldsymbol{p_k})=0$ otherwise.
Under this condition, the IMLS iterative scheme near $\boldsymbol{p}_i$ becomes 
\begin{equation}
\label{eq::g::iterative:continuous}
    \boldsymbol{f}_\text{IMLS}(\boldsymbol{q})=\left\langle\boldsymbol{q}-\boldsymbol{p}_i, \frac{\nabla \boldsymbol{{f}_\text{MLP}(\boldsymbol{p}_i)}}{\left\| \nabla \boldsymbol{{f}_\text{MLP}(\boldsymbol{p}_i)} \right\|} \right\rangle.
\end{equation}
Obviously, we have 
\begin{equation}
\boldsymbol{f}_\text{IMLS}(\boldsymbol{p}_i)=0, \;\;
\|\nabla_{\boldsymbol{q}} \boldsymbol{f}_\text{IMLS}\| = \left\| \frac{\nabla \boldsymbol{{f}_\text{MLP}(\boldsymbol{p}_i)}}{\left\| \nabla \boldsymbol{{f}_\text{MLP}(\boldsymbol{p}_i)} \right\|} \right\|=1,
\end{equation}
which implies that the IMLS surface tends to interpolate the given point cloud and approximates a signed distance field near the surface.
At the same time, by minimizing the loss, i.e., the difference between $\boldsymbol{f}_\text{MLP}$ and $\boldsymbol{f}_\text{IMLS}$, they tend to coincide with each other.
\end{proof}

\begin{figure*}[!htp]
    \centering
    \includegraphics[
      width=\textwidth
    ]{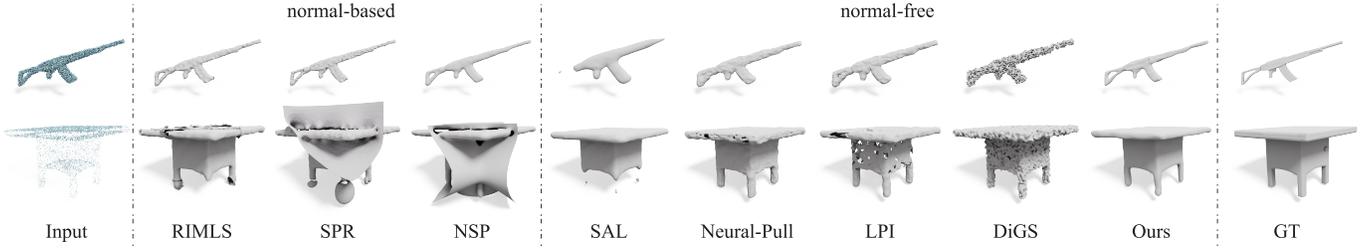}
    \vspace{-8mm}
    \caption{Qualitative comparison under ShapeNet~\cite{chang2015shapenet}. 
    All methods test on 10k noisy points and the oriented normals estimated by Dipole~\cite{metzer2021orienting} for normal-based methods.}
    \label{fig:shapenet}
\end{figure*}

\vspace{.5mm}
\noindent\textbf{Remark.}
We must point out that in real situations, 
the points are noisy and sparse, it requires a number of iterations to consistently transform $\boldsymbol{f}_\text{IMLS}$ or $\boldsymbol{f}_\text{MLP}$ into a real signed distance function.

It's also worth noting that the difference between $\boldsymbol{f}_\text{IMLS}$ and $\boldsymbol{f}_\text{MLP}$
is measured w.r.t~$\boldsymbol{Q}$, a pre-computed sample set nearby~$\boldsymbol{P}$.
Because of this, the values of $\boldsymbol{f}_\text{IMLS}$ and $\boldsymbol{f}_\text{MLP}$
are close to each other for points in~$\boldsymbol{Q}$,
rather than all points in~$\mathbb{R}^3$.
Their distance values are not necessarily identical to each other at points far from the input point set~$\boldsymbol{P}$.
An interesting fact is that $\boldsymbol{f}_\text{IMLS}$ fits the point cloud mainly on local information, but $\boldsymbol{f}_\text{MLP}$ fits the point cloud in a more global manner since its basis functions of $\boldsymbol{f}_\text{MLP}$ do not vanish even when~$\boldsymbol{q}$ is located infinitely far away. 
Therefore, when the input point cloud has missing parts,
$\boldsymbol{f}_\text{MLP}$ still can complete completing a reasonable shape,
but $\boldsymbol{f}_\text{IMLS}$ cannot.

\begin{table}[!htp]
\centering
\caption{\textbf{ShapeNet.} Quantitative results for surface reconstruction from
unoriented point clouds on the ShapeNet.
The methods test on 10k noisy points. 
RIMLS~\cite{oztireli2009RIMLS}, SPR~\cite{kazhdan2013screened}, and NSP~\cite{williams2021nsp} leverage oriented normals estimated by Dipole~\cite{metzer2021orienting}.}
\label{tab:shapenet}
\begin{tabular}{p{0.12\textwidth}|>{\centering\arraybackslash}p{0.09\textwidth}>{\centering\arraybackslash}p{0.07\textwidth}>{\centering\arraybackslash}p{0.07\textwidth}} 
\toprule
Method      & CD ($\times 100$)$\downarrow$ & NC $\uparrow$ & FS $\uparrow$  \\ 
\midrule
RIMLS~\cite{oztireli2009RIMLS}       & 1.04            & 0.85          & 0.64           \\
SPR~\cite{kazhdan2013screened}         & 2.36            & 0.81          & 0.57           \\
NSP~\cite{williams2021nsp}         & 1.60            & 0.86          & 0.61           \\
\midrule
SAL~\cite{SAL}        & 2.07            & 0.84          & 0.52           \\
Neural-Pull~\cite{NeuralPull} & 1.21            & 0.89          & 0.60           \\
LPI~\cite{LPI}         & 1.15            & 0.89          & 0.64           \\
DiGS~\cite{ben2022digs}       & 1.02            & 0.63          & 0.61           \\
\textbf{Ours}        & \textbf{0.64}   & \textbf{0.91} & \textbf{0.81}  \\
\bottomrule
\end{tabular}
\end{table}

\section{Experiments}
\label{sec:exp}
\subsection{Setup}

\vspace{.5mm}
\noindent
\textbf{Parameters.}
Note that we have a parameter~$r$
to define the ball for finding the point subset in~$\boldsymbol{P}$
for a given query point~$\boldsymbol{q}$.
We set the default search radius $r$ to be~$0.01 d_{\boldsymbol{P}}$, where $d_{\boldsymbol{P}}$ is the diagonal length of the bounding box of the input point cloud~$\boldsymbol{P}$.
If $\boldsymbol{B}_{r}(\boldsymbol{q})$ is empty,
we deem $\boldsymbol{q}$ as an invalid query point and simply ignore $\boldsymbol{q}$.
If $\left|\boldsymbol{B}_{r}(\boldsymbol{q})\right|$ is less than~$50$,
then we augment it to 50 points with a padding technique.
Otherwise, we downsample $\boldsymbol{B}_{r}(\boldsymbol{q})$ to 50 points.
We set~$\sigma_\text{IMLS} = \sqrt{d_{\boldsymbol{B}_{r}(\boldsymbol{q})}/ \left|\boldsymbol{B}_{r}(\boldsymbol{q})\right|}$ empirically following~\cite{huang2009consolidation}, where $d_{\boldsymbol{B}_{r}(\boldsymbol{q})}$ is the diagonal length of the bounding box of $\boldsymbol{B}_{r}(\boldsymbol{q})$. It's worth noting that $d_{\boldsymbol{B}_{r}(\boldsymbol{q})}$ depends on the  distribution of the points in~$\boldsymbol{B}_{r}(\boldsymbol{q})$, and thus cannot be directly determined by $r$. 
Besides, $\sigma_\text{coherence}$ is set to~$0.3$ by default.

\vspace{.5mm}
\noindent
\textbf{Dataset.} 
According to the size of the reconstruction target,
the task of surface reconstruction
can be divided into object-level and scene-level. 
For the object-level surface reconstruction, we evaluate our method on a total of five datasets, including Surface Reconstruction Benchmark (SRB)~\cite{berger2013benchmark}, ShapeNet~\cite{chang2015shapenet}, ABC~\cite{koch2019abc}, FAMOUS~\cite{erler2020points2surf}, and Thingi10K~\cite{zhou2016thingi10k}.
We also test our method on a real-scan object-level benchmark from~\cite{huang2022survey}.
For the scene-level surface reconstruction,
we conduct the experiments on the synthetic indoor scene dataset proposed by~\cite{peng2020convolutional}. 
All the output meshes are extracted with the improved Marching Cubes~\cite{lewiner2003efficient}.
The resolution for running marching cubes
is set to $512^3$ on the three datasets, including SRB, real-scanned object-level benchmark, and synthetic indoor scene dataset.
For the other datasets, including FAMOUS, ABC, and Thingi10K,
we use a resolution of~$256^3$, following Points2Surf~\cite{erler2020points2surf}.
The resolution of marching cubes for the ShapeNet dataset is also set to~$256^3$.

\vspace{.5mm}
\noindent
\textbf{Metrics.} 
As various approaches differ in evaluation metrics,
we have to tune our experimental settings to ensure a fair comparison.
For the SRB dataset, we follow DiGS~\cite{ben2022digs} to evaluate the related approaches with the Chamfer Distance (CD) and Hausdorff Distance (HD). 
The number of test points for computing the indicators is set to $10^6$.
For ShapeNet and the synthetic indoor scene dataset, we 
use three indicators, including Chamfer Distance (CD), normal consistency (NC), and F-Score (FS) with threshold values of 0.01, and 0.02 respectively. The number of test points is set to $10^5$.
For the ABC, FAMOUS, and Thingi10K datasets,
we make comparisons based on Chamfer Distance\footnote{Points2Surf implements different Chamfer Distance, which is not inconsistent with Ep. 10 in its main paper. See Appendix~B for details.},
in the same setting as Points2Surf~\cite{erler2020points2surf}.
The number of test points is set to $10^4$.
For the real-scan benchmark, we follow~\cite{huang2022survey} to report Chamfer Distance (CD), F-Score (FS), normal consistency (NC), and Neural Feature Similarity (NFS).
It's worth noting that the threshold of
the FS indicator is increased to 0.5 for real-scan data, and the number of test points is set
to $2\times10^5$.





\begin{table*}[!htp]
\centering
\caption{\textbf{ABC, Famous, Thingi10k.} Chamfer distance $\times 100$ on ABC, Famous, and Thingi10k test sets with variable Gaussian noise ($\sigma$ uniformly picked in $[0,0.05L]$, $L$ largest box length), as prepared by~\cite{erler2020points2surf}: 'no-n.' (no noise), 'var-n.' (variable noise, $\sigma$ in $[0,0.05L]$), 'med-n.' ( $\sigma=0.01 L)$, 'max-n.' $(\sigma=0.05 L)$, 'sparse' (5 scans with $\sigma=0.01L$), 'dense' (30 scans with $\sigma=0.01L$). Only SPR uses normals.}
\label{tab:quantitative_p2s}
\resizebox{\linewidth}{!}{%
\begin{tabular}{l|c|ccc|ccccc|ccccc} 
\toprule
              &                           & \multicolumn{3}{c|}{ABC}          & \multicolumn{5}{c|}{FAMOUS}                                        & \multicolumn{5}{c}{Thingi10k}                                     \\
Method~       & Normals or Supervision       & no-n.         & var-n.        & max-n.        & no-n.         & med-n.        & max-n.        & sparse        & dense         & no-n.         & med-n.        & max-n.        & sparse        & dense          \\ 
\midrule
DeepSDF~\cite{park2019deepsdf}      & \checkmark & 8.41          & 12.51         & 11.34         & 10.08         & 9.89          & 13.17         & 10.41         & 9.49          & 9.16          & 8.83          & 12.28         & 9.56          & 8.35           \\
AatlasNet~\cite{groueix2018Aatlas}    & \checkmark & 4.69          & 4.04          & 4.47          & 4.69          & 4.54          & 4.14          & 4.91          & 4.35          & 5.29          & 5.19          & 4.90          & 5.64          & 5.02           \\
SPR~\cite{kazhdan2013screened}          & \checkmark & 2.49          & 3.29          & 3.89          & 1.67          & 1.80          & 3.41          & 2.17          & 1.60          & 1.78          & 1.81          & 3.23          & 2.35          & 1.57           \\
Points2Surf~\cite{erler2020points2surf}  & \checkmark & 1.80          & 2.14          & 2.76          & 1.41          & 1.51          & \textbf{2.52} & 1.93          & \textbf{1.33} & 1.41          & 1.47          & 2.62          & 2.11          & \textbf{1.35}  \\
POCO~\cite{Boulch2022poco}   & \checkmark & \textbf{1.70} & \textbf{2.01} & \textbf{2.50} & \textbf{1.34} & \textbf{1.50} & 2.75          & \textbf{1.89} & 1.50          & \textbf{1.35} & \textbf{1.44} & \textbf{2.34} & \textbf{1.95} & 1.38           \\ 
\midrule
SAL~\cite{SAL}          & $\times$                  & 3.44          & 4.77          & 7.10          & 3.88          & 4.16          & 9.92          & 4.93          & 3.53          & 2.85          & 3.22          & 7.80          & 4.03          & 2.88           \\
Neural-Pull~\cite{NeuralPull}  & $\times$                  & 3.62          & 6.33          & 6.36          & 2.66          & 3.17          & 5.96          & 3.28          & 3.65          & 1.86          & 2.45          & 5.82          & 2.92          & 2.56           \\
DiGS~\cite{ben2022digs}         & $\times$                  & 2.39          & 2.99          & 5.14          & 1.61          & 2.42          & 4.92          & 2.58          & 2.09          & 1.59          & 2.74          & 5.89          & 3.18          & 2.48           \\
\textbf{Ours} & $\times$                  &   \textbf{1.69}            &   \textbf{2.49}            &     \textbf{5.09}          & \textbf{1.34} & \textbf{1.62} &    \textbf{4.71}           &    \textbf{2.24}           &     \textbf{1.48}          &   \textbf{1.41}            &   \textbf{1.69}            &      \textbf{5.53}         &     \textbf{2.10}          &   \textbf{1.53}             \\ 
\bottomrule\end{tabular}
}
\end{table*}
\begin{table}[!htp]
\centering
\caption{
\textbf{Comparison on object-level real scans from~\cite{huang2022survey}.}
The methods using surface normals only during training
 are annotated with $*$.
}

\label{tab:real_scan}

\resizebox{\linewidth}{!}{
\begin{tabular}{l|cc|cccc} 
\toprule
Method       & Normals                   & Supervision               & CD ($\times 100 $) $\downarrow$ & FS ($\times 100$) $\uparrow$   & NC ($\times 100$) $\uparrow$ & NFS ($\times 100$) $\uparrow$  \\ 
\midrule
BPA~\cite{bernardini1999ball}           &        \checkmark                   & $\times$                  & 40.37              & 80.95          & 87.56               & 68.69                \\
SPR~\cite{kazhdan2013screened}           &      \checkmark                      & $\times$                  & \textbf{31.05}     & \textbf{87.74} & 94.94               & \textbf{89.38}       \\
RIMLS~\cite{oztireli2009RIMLS}         &       \checkmark                     & $\times$                  & 32.80               & 87.05          & 91.97               & 85.19                \\
IGR~\cite{IGR}           &       \checkmark                     & $\times$ & 32.70              & 87.18          & \textbf{95.99}      & 89.10                \\ 
\midrule
OccNet~\cite{mescheder2019onet}        &          \checkmark                  &       \checkmark                    & 232.71             & 17.11          & 80.96               & 39.70                 \\
DeepSDF~\cite{park2019deepsdf}       &          \checkmark                  &     \checkmark                      & 263.92             & 19.83          & 77.95               & 40.95                \\
LIG~\cite{jiang2020local}           &           \checkmark                 &     \checkmark                      & 48.75              & 83.76          & 92.57               & 81.48                \\
ParseNet~\cite{sharma2020parsenet}      &           \checkmark                 &    \checkmark                       & 149.96             & 38.92          & 81.51               & 45.67                \\
Points2Surf~\cite{erler2020points2surf}   & $*$                       &    \checkmark                       & 48.93              & 80.89          & 89.52               & 81.83                \\
DSE~\cite{rakotosaona2021learning}           &  $\times$                  &   \checkmark                        & \textbf{32.16}     & \textbf{86.88} & 87.20                & 76.81                \\
DeepMLS~\cite{Liu2021MLS}       & $*$                       &  \checkmark                         & 38.46              & 82.44          & 93.31               & \textbf{85.30}       \\
POCO~\cite{Boulch2022poco}          & $\times$     & \checkmark     & 56.53              & 83.02          & \textbf{93.68}      & 84.05                \\ 
\midrule
GD~\cite{cohen2004greedy}            & $\times$ & $\times$ & 31.72              & 87.51          & 88.86               & 82.20                \\
SALD~\cite{atzmon2021sald}          & $\times$                  & $\times$                  & \textbf{31.13}     & \textbf{87.72} & 94.68               & 86.86                \\
DiGS~\cite{ben2022digs}          & $\times$                  & $\times$                  & 70.23              & 80.64          & 93.28               & 77.25                \\
PCP~\cite{ma2022surface}           & $\times$                  & $\times$                  & 71.66              & 54.01          & 94.96               & 77.20                \\
\textbf{Ours} & $\times$                  & $\times$                  & 34.96              & 86.69          & \textbf{95.04}      & \textbf{88.69}       \\
\bottomrule
\end{tabular}
}
\end{table}

\subsection{Object-level Reconstruction}

\subsubsection{Surface Reconstruction Benchmark (SRB)}
The SRB contains five shapes with complex geometry and topology,
whose noisy simulated scans are provided by Williams et al.~\cite{williams2019dgp}. 
The difficulties of reconstruction lie in the high genus,
rich details, missing data, and varying feature sizes.
We report the average scores in terms of the Chamfer distance and the Hausdorff distance in Table~\ref{table:srb}. 
The methods for comparison include SAL~\cite{SAL}, IGR~\cite{IGR}, SIREN~\cite{sitzmann2020siren}, PHASE~\cite{lipman2021phase}, Shape As Points (SAP)~\cite{peng2021shape}, iPSR~\cite{hou2022iterative}, DiGS~\cite{ben2022digs}, and PredictableContextPrior (PCP)~\cite{ma2022surface} since they do not require extra supervision or normals. 
We also test IGR and PHASE using Fourier features~\cite{tancik2020fourier}
under the mode of being free of normals,
which is named IGR+FF and PHASE+FF, respectively.
From Table~,\ref{table:srb} we can see that our method outperforms the SOTA methods. 

We visualize qualitative results in Fig.~\ref{fig:srb}. 
As the extension of Poisson reconstruction, SAP and iPSR suffer from 
shallow gap issues that wrongly close gaps.
The resulting surfaces produced by PCP have topological errors 
and miss geometric details. 
DiGS is comparable to ours on the dataset 
except that DiGS is weak in reconstructing sharp or neat structures
(see the first row of Fig.~\ref{fig:srb}).

\subsubsection{ShapeNet}
The ShapeNet~\cite{chang2015shapenet} dataset contains CAD models with diverse shapes.
We use 13 categories of shapes in ShapeNet to conduct experiments,
in the same preprocessing as Williams et al.~\cite{williams2021nsp}.
For each mesh, we randomly sample 10k non-uniformly distributed points
and add Gaussian noise with a standard deviation of $0.005$.
The  methods included for comparison are RIMLS~\cite{oztireli2009RIMLS}, Screen Poisson Surface Reconstruction (SPR)~\cite{kazhdan2013screened}, Neural Splines (NSP)~\cite{williams2021nsp}, SAL~\cite{SAL}, Neural-Pull~\cite{NeuralPull}, DiGS~\cite{ben2022digs}, and Latent Partition Implicit (LPI)~\cite{LPI}.
It's worth noting that RIMLS, SPR, and NSP need oriented normals,
we use Dipole~\cite{metzer2021orienting} to equip the points with normals. 
As pointed out in~\cite{metzer2021orienting}, the approach of computing normals is robust to thin structures and noisy inputs.
For better dealing with noise, 
we use a larger local-size parameter $r=0.03 d_{\boldsymbol{P}}$.
In this situation, the standard number of points in $\boldsymbol{B}_{r}$
is set to 100. 
The statistics are reported in Table~,\ref{tab:shapenet} and we provide visual results in Fig.~\ref{fig:shapenet}.
Although Dipole~\cite{metzer2021orienting} is strong in estimating reliable oriented normals,
RIMLS, SPR, and NSP still cannot produce high-quality results
since it is hard to infer reliable normals when the input shape contains thin structures
and the corresponding point cloud is noisy. 
SAL is noise-resistant to some degree, but it produces over-smooth results and has trouble capturing thin strictures.
Noise affects Neural-Pull, LPI, and DiGS, causing unwanted surface changes and unnecessary topological holes.
Among them, DiGS has the lowest NC scores. 
In contrast, our method is superior to the above methods in recovering the geometric details with the help of the mutual regulation between MLP and IMLS.

\begin{figure*}[!htp]
    \centering
    \includegraphics[
      width=\textwidth
    ]{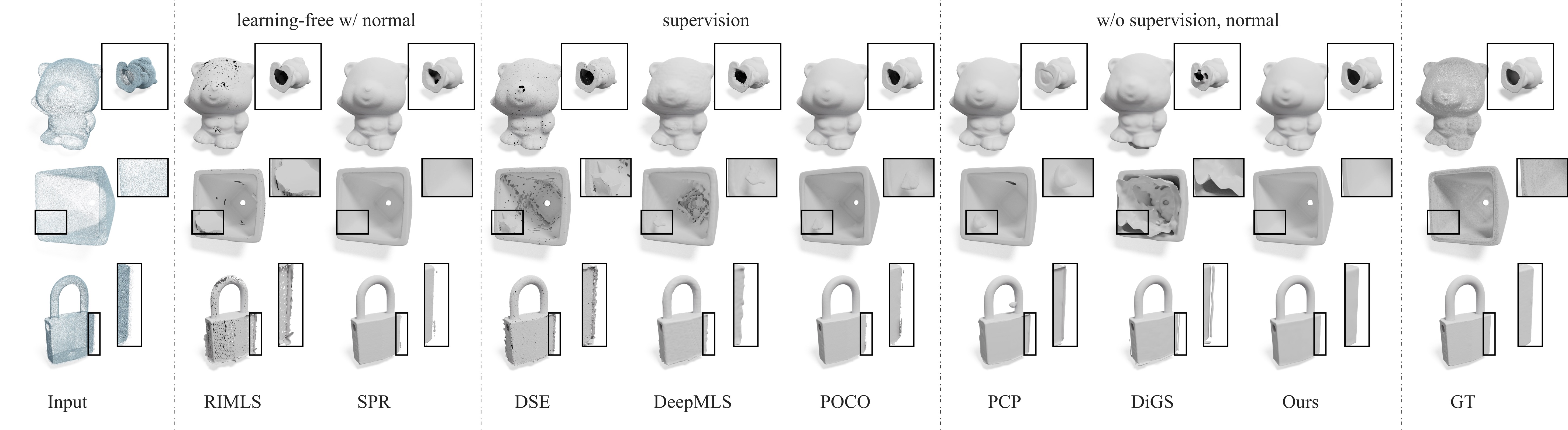}
    \vspace{-8mm}
    \caption{Qualitative comparison under real scans from~\cite{huang2022survey}.
    Our method is more capable than others of handling thin structures, concave parts, and misalignment input.
    }
    \label{fig:real20}
\end{figure*}

\subsubsection{ABC, FAMOUS, and Thingi10K}
In this section, we conduct experiments under the three datasets that are preprocessed into points by Points2Surf~\cite{erler2020points2surf}.
It applies BlenSor~\cite{gschwandtner2011blensor} to simulate real scanners to produce ray-direction noise. 
Each shape is scanned by 10 virtual cameras with variable Gaussian noise and variable point densities. 
Let $\sigma$ be the standard deviation of scanned points. 
We use four different noise-level settings: $\sigma = 0$ (no noise, no-n), $\sigma = 0.01L$ (medium noise, med-n), $\sigma \in [0, 0.05L]$ (variable noise, var-n), and $\sigma = 0.05L$ (maximum noise, max-n),  where $L$ is the length of the longest bounding box edge.
Moreover, there are ``sparser'' (5 scans with $\sigma = 0.01L$) and ``dense'' (30 scans with $\sigma = 0.01L$) variants to test the robustness of the point density.
We adjust the local-size parameter to adapt to different noise levels:
$r=0.01 d_{\boldsymbol{P}}, \left|\boldsymbol{B}_{r}(\boldsymbol{q}_j)\right|=50$ (no-n)
$r=0.03 d_{\boldsymbol{P}}, \left|\boldsymbol{B}_{r}(\boldsymbol{q}_j)\right|=100$ (sparse, dense, med-n),
$0.1 d_{\boldsymbol{P}}, \left|\boldsymbol{B}_{r}(\boldsymbol{q}_j)\right|=200$ (var-n, max-n).

We compare our method with state-of-the-art data-driven surface reconstruction methods, including DeepSDF~\cite{park2019deepsdf}, AtlasNet~\cite{groueix2018Aatlas},  Points2Surf~\cite{erler2020points2surf}, and POCO~\cite{Boulch2022poco}.
We also include Screened Poisson Surface Reconstruction (SPR)~\cite{kazhdan2013screened}
for comparison. Note that SPR needs oriented normals,
and thus we employ PCP-Net~\cite{PCP} for estimating normal vectors to facilitate the execution of screened Poisson.
We also compare our method with the overfitting methods, including SAL~\cite{SAL}, Neural-Pull~\cite{NeuralPull}, and DiGS~\cite{ben2022digs}.
Table~\ref{tab:quantitative_p2s} reports the quantitative results under all variants of the dataset.
See Appendix~B.4 for qualitative results.
Both quantitative and qualitative results show that
our method is superior to the other overfitting methods across different noise-level datasets.
Similarly, SAL produces over-smooth surfaces, possibly with topological errors.
Neural-Pull and DiGS are sensitive to noise,
and thus the presence of noise makes them produce non-smooth or even topologically incorrect results. 
In fact, both of them assume that the input points are accurately on the 0-isosurface,
making those noisy points pull the surface toward an unwanted configuration. 
We must point out that our scores about Neural-Pull~\cite{NeuralPull} are different from the scores in their original publication
due to different formulations of Chamfer Distance; Check Appendix~B.4 for details.

\vspace{.5mm}
\noindent\textbf{Remark.}
Compared with the two latest supervision methods, i.e., Points2Surf and POCO,
our method can produce comparable or even better results for most datasets.
For example, quantitative statistics on the ABC no-noise dataset show that our method
outperforms Points2Surf and POCO; See Table~\ref{tab:quantitative_p2s}.
However, we also note that on those max-noise datasets, our strategy cannot currently outperform the supervision methods.


\subsubsection{Real Scans}
We also present an experiment on the real-scan benchmark proposed by~\cite{huang2022survey}.
This dataset contains 20 objects that are commonly seen in life.
The scan quality varies with different types of material.
We take the point clouds scanned by the consumer-grade depth camera SHINING Einscan SE as input and the point clouds scanned by a high-precision camera OKIO 5M as the ground truth.
The input point clouds have various artifacts, such as misaligned scans, noise, and missing parts.
Fig.~\ref{fig:real20} visualizes the reconstruction results.
The quantitative statistics are available in Table~\ref{tab:real_scan}.
RIMLS and DeepMLS, in spite of being based on the IMLS, 
struggle to separate the points on the opposite sides of a thin structure
and address misaligned scans.
Our method, by contrast, can better manifest thin structures 
even in the presence of noise and misalignment. 
Furthermore, our method has the best NFS scores compared with those methods 
that do not use normals or supervision. 
Note that our NFS scores are lower than SPR~\cite{kazhdan2013screened} and IGR~\cite{IGR},
but they utilize the normals given by the scanner. 

\begin{table}[!htp]
\centering
\caption{\textbf{Scene-level reconstruction on synthetic rooms.} Qualitative comparison for surface reconstruction from unoriented point clouds on the synthetic room indoor scene dataset provided by~\cite{peng2020convolutional}. }
\label{tab:room}

\begin{tabular}{p{0.12\textwidth}|>{\centering}p{0.09\textwidth}>{\centering}p{0.06\textwidth}>{\centering\arraybackslash}p{0.06\textwidth}} 
\toprule
Method     & CD ($\times 100$) $\downarrow$ & NC $\uparrow$ & FS $\uparrow$  \\ 
\midrule
SAL~\cite{SAL}         & 4.33         & 0.64        & 0.32                \\
IGR~\cite{IGR}        & 7.28         & 0.71        & 0.44                \\
Neural-Pull~\cite{IGR} & 1.18           & 0.87          & 0.87                  \\
PCP~\cite{ma2022surface}         & 1.12           & 0.89          & 0.91                  \\
DiGS~\cite{ben2022digs}        & \textbf{0.91}           & 0.68          & 0.82                  \\ 
\midrule
SA-CONet~\cite{tang2021sa}    & 1.48          & 0.89         & 0.86                 \\ 
\midrule
\textbf{Ours}        & \textbf{0.91}           & \textbf{0.90}          & \textbf{0.93}                  \\
\bottomrule
\end{tabular}
\end{table}

\begin{figure*}[!htp]
    \centering
    \includegraphics[
      width=\textwidth
    ]{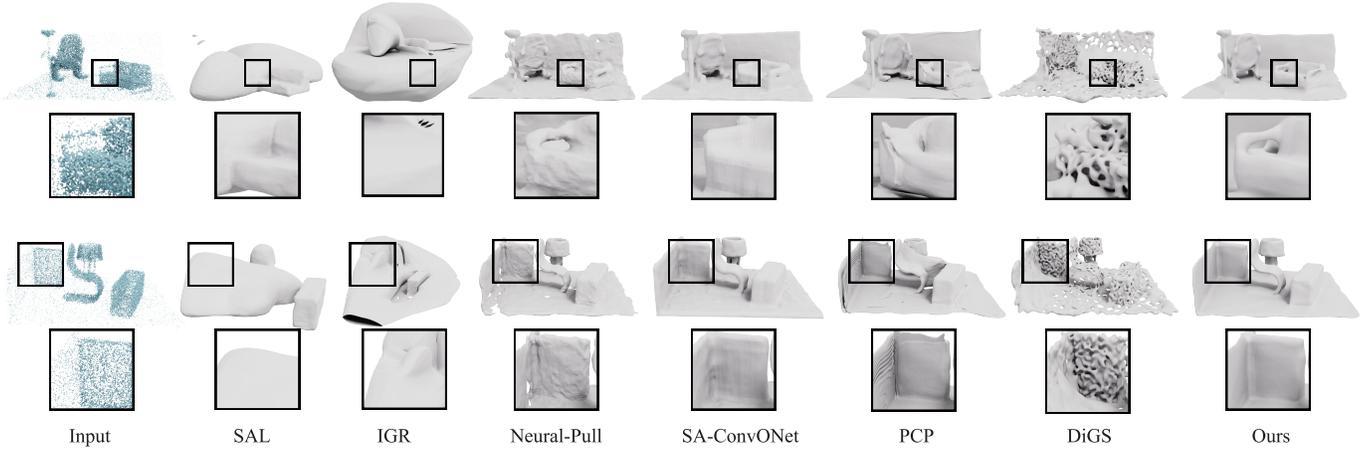}
    \vspace{-8mm}
    \caption{
    Scene-level reconstruction under synthetic-scan rooms~\cite{peng2020convolutional}. The input is 30k noisy points.
    }
    \label{fig:room}
\end{figure*}

\subsection{Scene-level Reconstruction}
To investigate whether our method possesses the scalability to indoor scene reconstruction, we further conduct experiments on the synthetic indoor scene dataset~\cite{peng2020convolutional}.
We use the split plan of~\cite{tang2021sa}.
Each scene has 30k points, and we add Gaussian noise with a standard deviation~$0.005$.
The compared methods include SAL~\cite{SAL}, IGR~\cite{IGR} (w.o. normal), Neural-Pull~\cite{NeuralPull}, SA-ConvONet~\cite{tang2021sa}, PredictableContextPrior (PCP)~\cite{ma2022surface}, and DiGS~\cite{ben2022digs}.
Note that SA-ConvONet is optimized on the basis of supervised ConvONet, while the others are overfitting without any supervision.
From the qualitative comparison shown in Fig.~\ref{fig:room}, we can
see that some holes in chairs and thin lamps can still be recovered
by our approach, while others cannot capture these small-sized geometric features. 
This shows that our method can deal with scene-level objects.
Statistics in Table~\ref{tab:room} demonstrate the superiority of our approach.





\subsection{Ablation Study and Discussion}
\label{ssec:ablation}

\begin{figure}[!htp]
    \centering
    \includegraphics[
      width=0.48\textwidth,
    ]{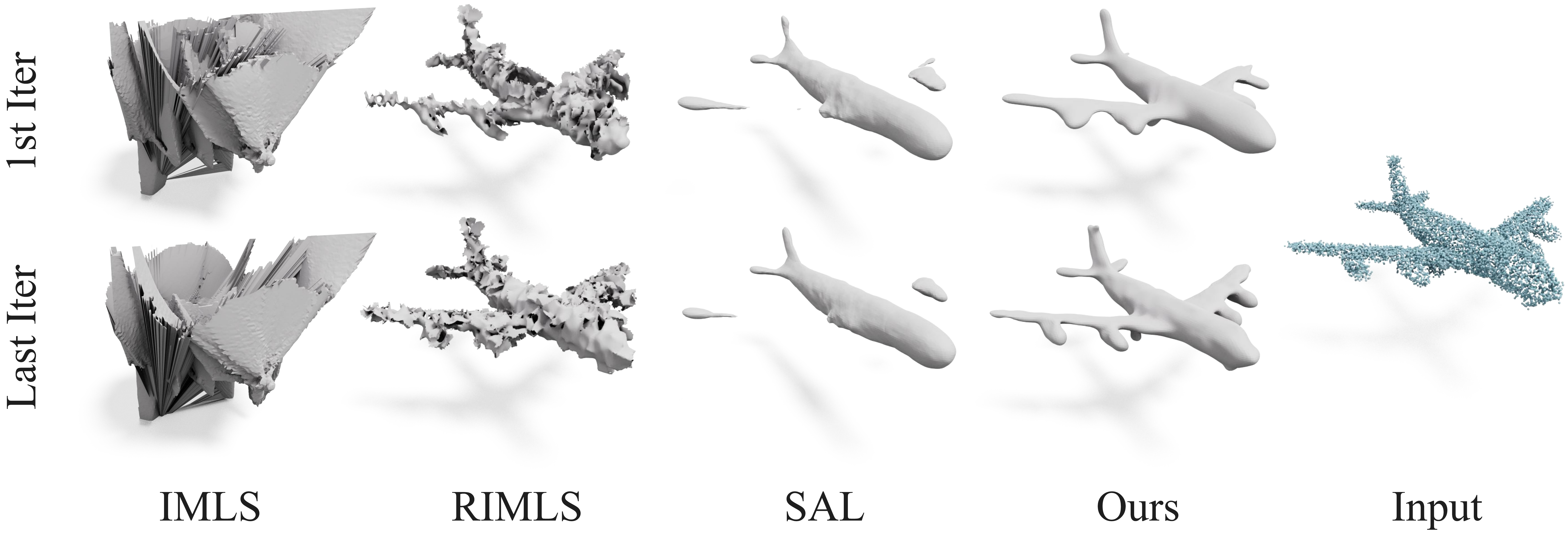}
    \vspace{-3mm}
    \caption{
    Ablation study about Mutual Regularization.
    }
    \label{fig:abl_mutal}
\end{figure}

\noindent
\textbf{Impact of Mutual Regularization.}
To conduct the ablation study,
we compare three different strategies:
(1)~IMLS-self-regularization: we modify the IMLS~\cite{kolluri2008imls} and the RIMLS~\cite{oztireli2009RIMLS} to develop an IMLS based self-regularization iterative scheme.
(2)~MLP-self-regularization: we first leverage SAL~\cite{SAL} to fit the shape and then use the trained model to supervise the new MLP network.
(3)~Mutual-regularization: using the MLP to provide normals to the IMLS while using the IMLS to provide distances to the MLP as the reference, which is the strategy proposed in this paper. 
We have a couple of observations.
First, all the strategies make the resulting surfaces change gradually, but within different iterations.
Also, the final surfaces at the convergence differ from each other. 
As reported in Tab.~\ref{tab:abl_mutal} and Fig.~\ref{fig:abl_mutal}, in contrast to the other two strategies,
the mutual-regularization strategy has a higher convergence rate, and the final surface is closer to the ground truth. 

\begin{table}[!htp]
\centering
\caption{\textbf{Ablation Study about Mutual Regularization.}}
\label{tab:abl_mutal}
\begin{tabular}{l|l|ccc} 
\toprule
                               &         & CD($\times100$)$\downarrow$ & NC$\uparrow$    & FS$\uparrow$    \\ 
\midrule
\multirow{2}{*}{IMLS~\cite{kolluri2008imls}}         & 1st iter  & 10.97    & 0.45   & 0.10  \\
                               & last iter & 10.84    & 0.46   & 0.10  \\ 
\midrule
\multirow{2}{*}{RIMLS~\cite{oztireli2009RIMLS}}        & 1st Iter  & 0.73     & 0.60  & 0.75  \\
                               & last Iter & 0.61     & 0.65~ & 0.82  \\ 
\midrule
\multirow{2}{*}{SAL~\cite{SAL}}  & 1st Iter  &  3.29        & 0.74      &  0.35     \\
                               & last Iter & 2.72         &  0.77     &  0.42     \\ 
\midrule
\multirow{2}{*}{\textbf{Ours}} & 1st Iter  & 1.09     & 0.85  & 0.59  \\
                               & last Iter & 0.45     & 0.92  & 0.94  \\
\bottomrule
\end{tabular}
\end{table}

\begin{figure}[!htp]
    \centering
    \includegraphics[
      width=0.45\textwidth,
    ]{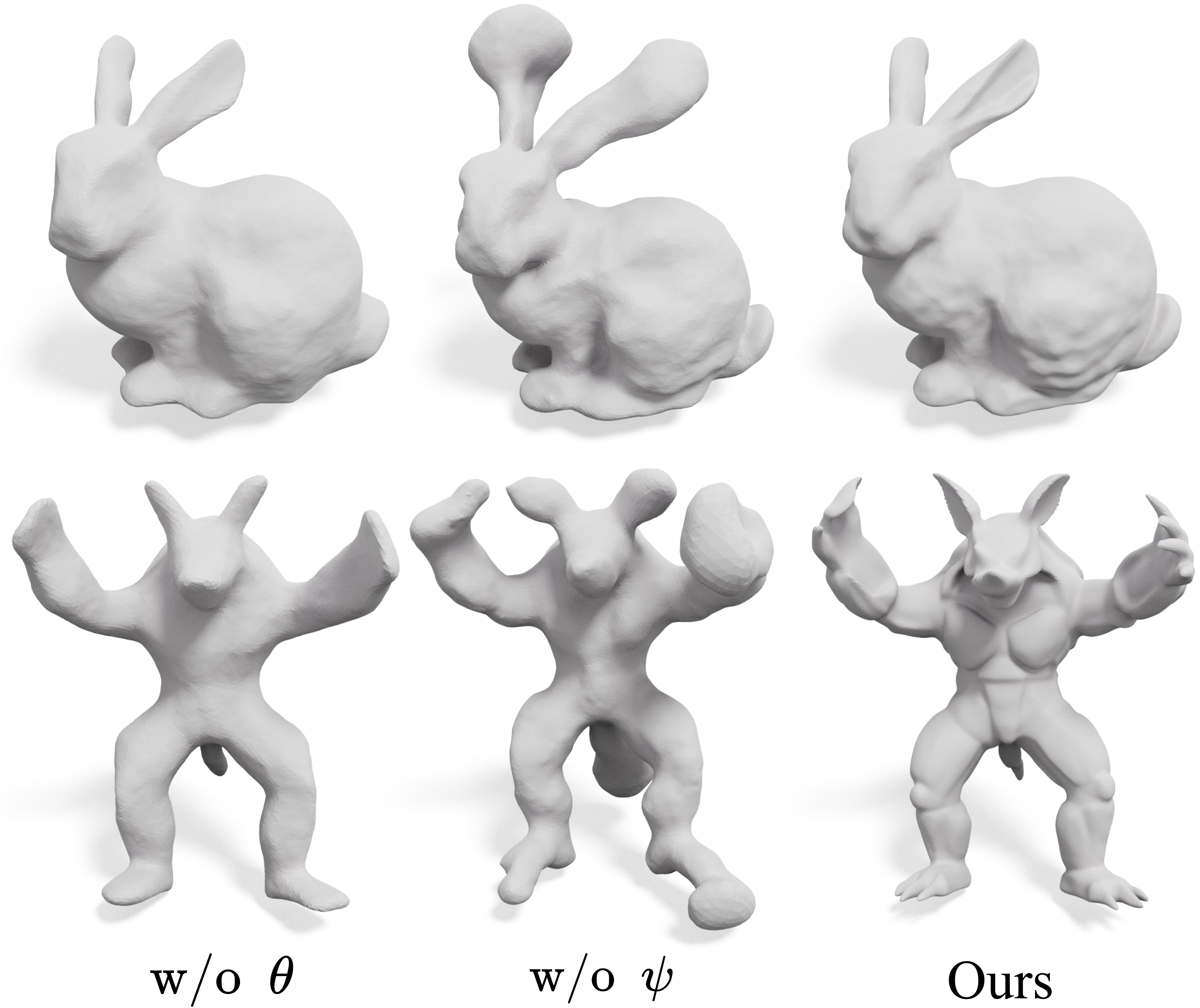}
    \vspace{-3mm}
    \caption{
    Reconstruction results without using distance weighting term $\theta$ (left), without using gradient weighting term $\psi$ (middle), and vanilla including both (right).
    }
    \label{fig:abl_term}
\end{figure}

\noindent
\textbf{The Impact of Loss Terms.} 
By switching off the gradient weighting scheme~$\psi$ and the distance weighting scheme~$\theta$ respectively, we
come to observe how different the reconstructed results become. 
We use ``w/o gradient'' and ``w/o distance'' to denote the options. 
The qualitative results and the quantitative results are shown in Fig.~\ref{fig:abl_term}
and Table.~\ref{tab:abl_lossterm}, respectively. 
It shows that without $\psi$ or $\theta$, the overall accuracy decreases.
Under the mode of ``w/o gradient'',
it is hard to enforce the spatial coherence of gradients,
causing conspicuous artifacts around thin structures or sharp tips. 
Under the mode of ``w/o distance'', 
the ability to preserve geometric details
is weakened due to the fact that
$\theta$ emphasizes more on the contribution of a local part. 

\begin{figure}[!htp]
    \centering
    \includegraphics[
      width=0.48\textwidth,
    ]{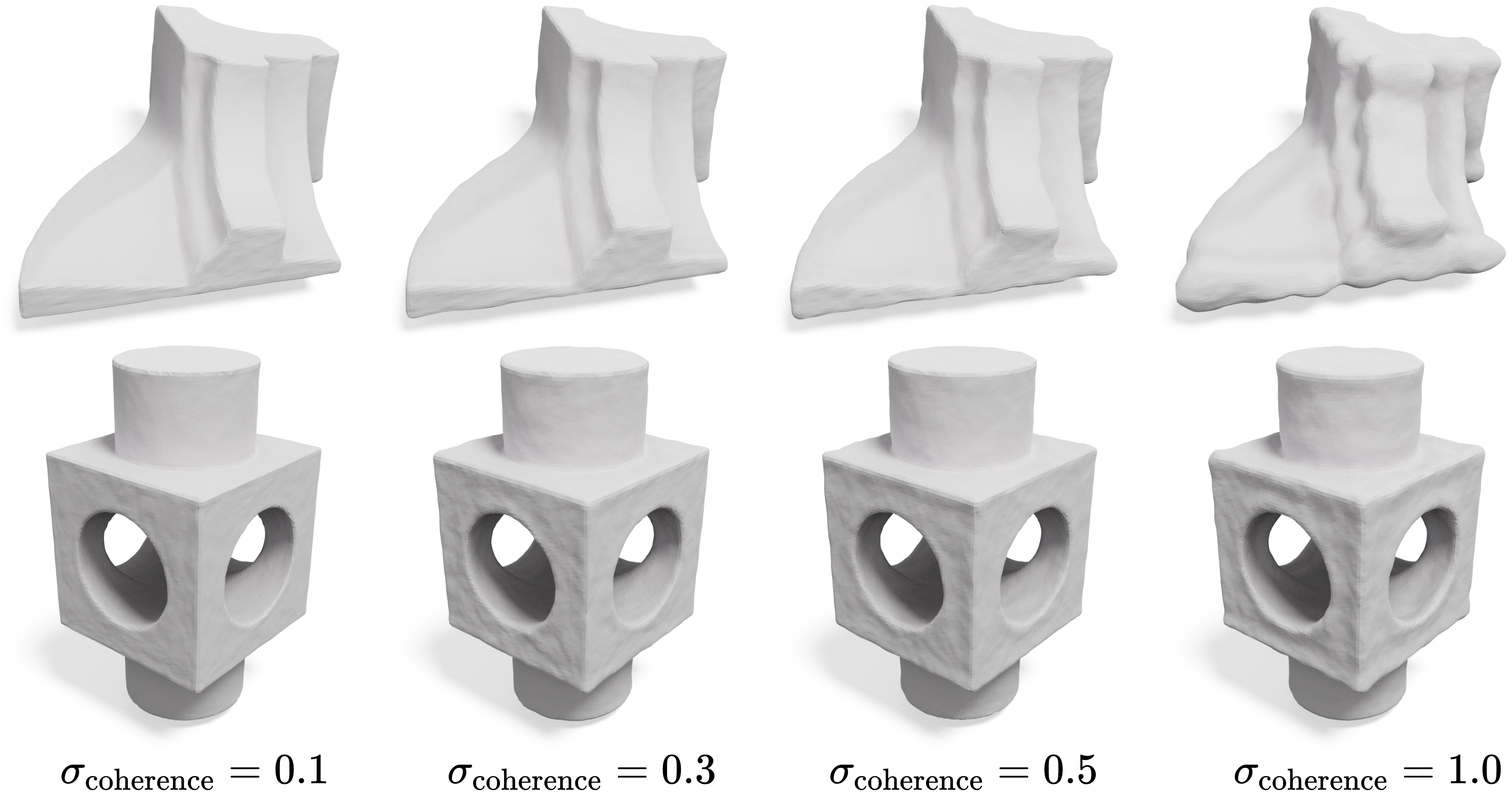}
    \vspace{-3mm}
    \caption{
    Visual comparison of different $\sigma_\text{coherence}$'s. A small $\sigma_\text{coherence}$ tends to better manifest sharp features.
    }
    \label{fig:sigma_n}
\end{figure}

\vspace{.5mm}
\noindent
\textbf{The Impact of Parameter $\sigma_\text{coherence}$.} 
As mentioned above, $\psi$ is to enforce the spatial coherence of gradients.
The parameter $\sigma_\text{coherence}$ in $\psi$ 
is used to tune the degree to which sharp features are preserved.
As shown in Fig.~\ref{fig:sigma_n},
a small $\sigma_\text{coherence}$ tends to better manifest sharp features,
while a large $\sigma_\text{coherence}$ tends to reconstruct a smoother surface.
In the default setting, we set $\sigma_\text{coherence}=0.3$ to balance the ability
of feature preserving and noise resistance.
The quantitative results are shown in Table.~\ref{tab:abl_lossterm}.
%

\begin{table}[!htp]
\centering
\caption{\textbf{Ablation study of loss term.} We show the Chamfer Distance scores with different options.}
\label{tab:abl_lossterm}
\begin{tabular}{l|cc} 
\toprule
                 & \multicolumn{2}{c}{FAMOUS}      \\
                 & no-n.         & med-n.          \\ 
\midrule
w/o $\theta$     & 1.83          & 2.13            \\
w/o $\psi$       & 2.01          & 2.17            \\
\textbf{vanilla} & \textbf{1.34} & \textbf{1.62}  \\ 
\midrule
$\sigma_\text{coherence}=0.1$   & \textbf{1.34}          & 1.68            \\
$\sigma_\text{coherence}=0.3$   & \textbf{1.34} & \textbf{1.62}   \\
$\sigma_\text{coherence}=0.5$   & 1.40          & 1.71            \\
$\sigma_\text{coherence}=1.0$   & 1.79          & 1.75            \\
\bottomrule
\end{tabular}
\end{table}

\begin{figure}[!htp]
    \centering
    \includegraphics[
      width=0.48\textwidth,
    ]{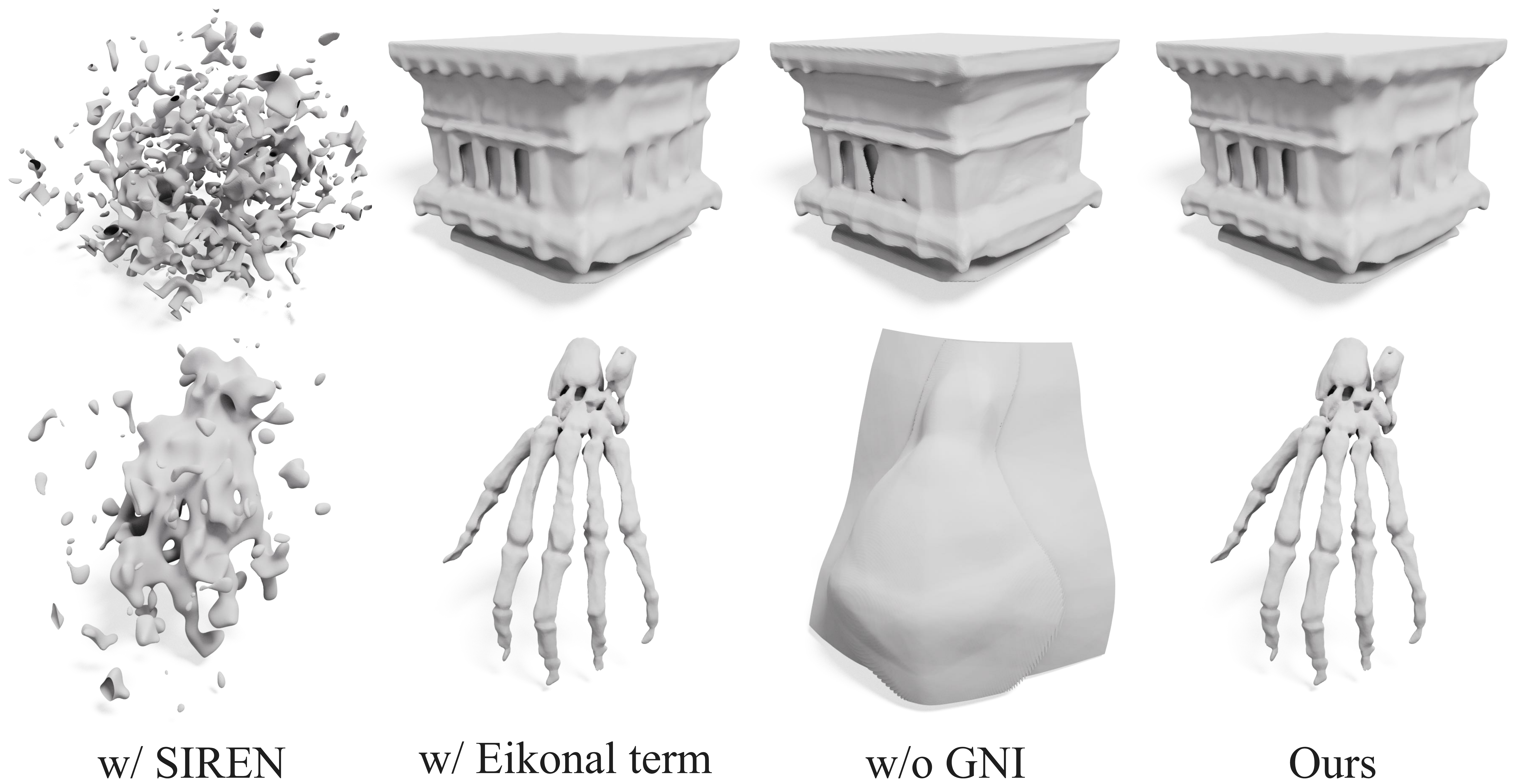}
    \vspace{-3mm}
    \caption{
    Ablation studies of network components.
    }
    \label{fig:abl_network}
\end{figure}

\vspace{.5mm}
\noindent
\textbf{Impact of Network Components.}
We explore the possibilities of improving the performance using some known techniques. Specially, we come to discuss the influence of SIREN~\cite{sitzmann2020siren} (activation function), Eikonal term~\cite{IGR} (gradients constraints), and geometric network initialization~\cite{SAL} under the no-noise FAMOUS dataset.
The qualitative results and the quantitative results are shown in Fig.~\ref{fig:abl_network} and Table.~\ref{tab:net_comp}, respectively. 

First, we replace Softplus with SIREN~\cite{sitzmann2020siren} in terms of the activation function. We use ``w/ SIREN '' to denote this option but cannot see any performance improvement.
For example, under the option of ``w/ SIREN '', it may lead to the occurrence of outlier surfaces due to its non-monotonously.

Second, the Eikonal term is enforced in many existing research works. We used an option of ``w/ Eikonal term'' to explicitly add such a constraint but didn't observe any change. 
The reason lies in that the constraint about the gradient norm has been fully considered in our loss design. 

Finally, we use an option of ``w/o GNI'' to initialize the normals randomly.
Even so, our method can still converge to an SDF, and sometimes the SDF
can also manifest the real shape, which is due to the strong mutual regularization 
between the MLP and the IMLS.

\begin{table}
\centering
\caption{\textbf{Ablation Study about Network Components.} We show the Chamfer Distance scores under different options.}
\label{tab:net_comp}
\begin{tabular}{l|c} 
\toprule
                & FAMOUS no-n.   \\ 
\midrule
w/ SIREN        & 7.16           \\
w/ Eikonal term & \textbf{1.34}  \\
w/o GNI         & 3.48           \\
\textbf{vanilla}         & \textbf{1.34}  \\
\bottomrule
\end{tabular}
\end{table}

\begin{figure}[!htp]
    \centering
    \includegraphics[
      width=0.48\textwidth,
    ]{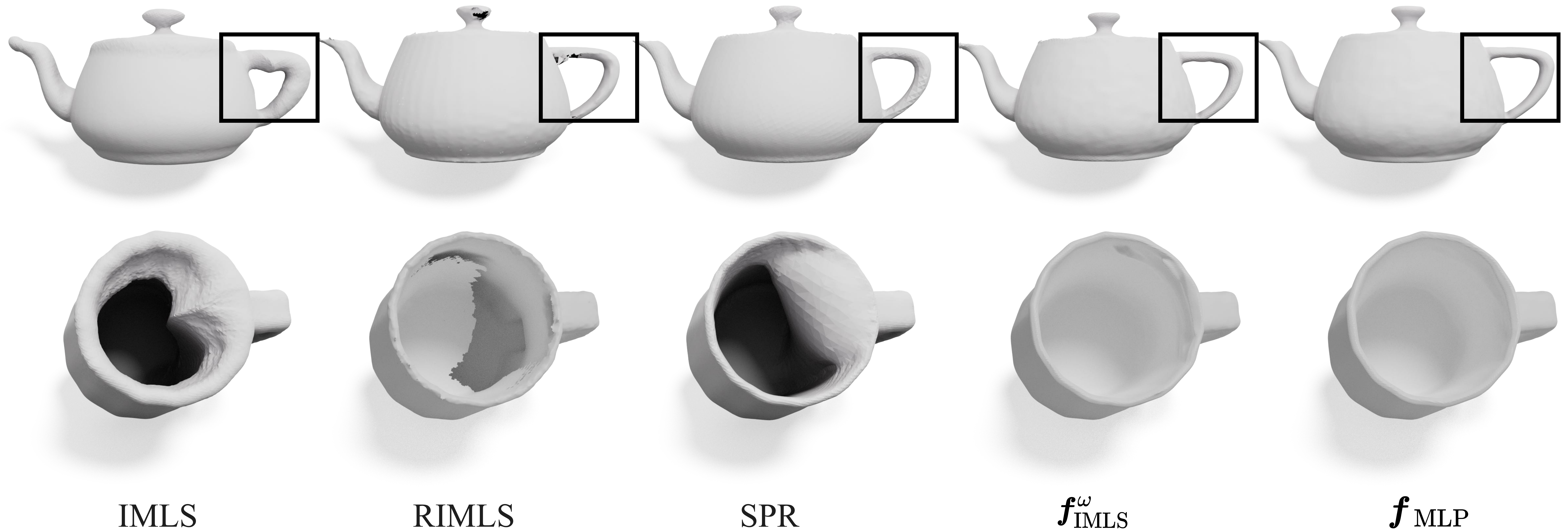}
    \vspace{-3mm}
    \caption{
    Comparison between different IMLS methods.
    }
    \label{fig:abl_imls}
\end{figure}

\begin{figure}[!htp]
    \centering
    \includegraphics[
      width=0.45\textwidth,
    ]{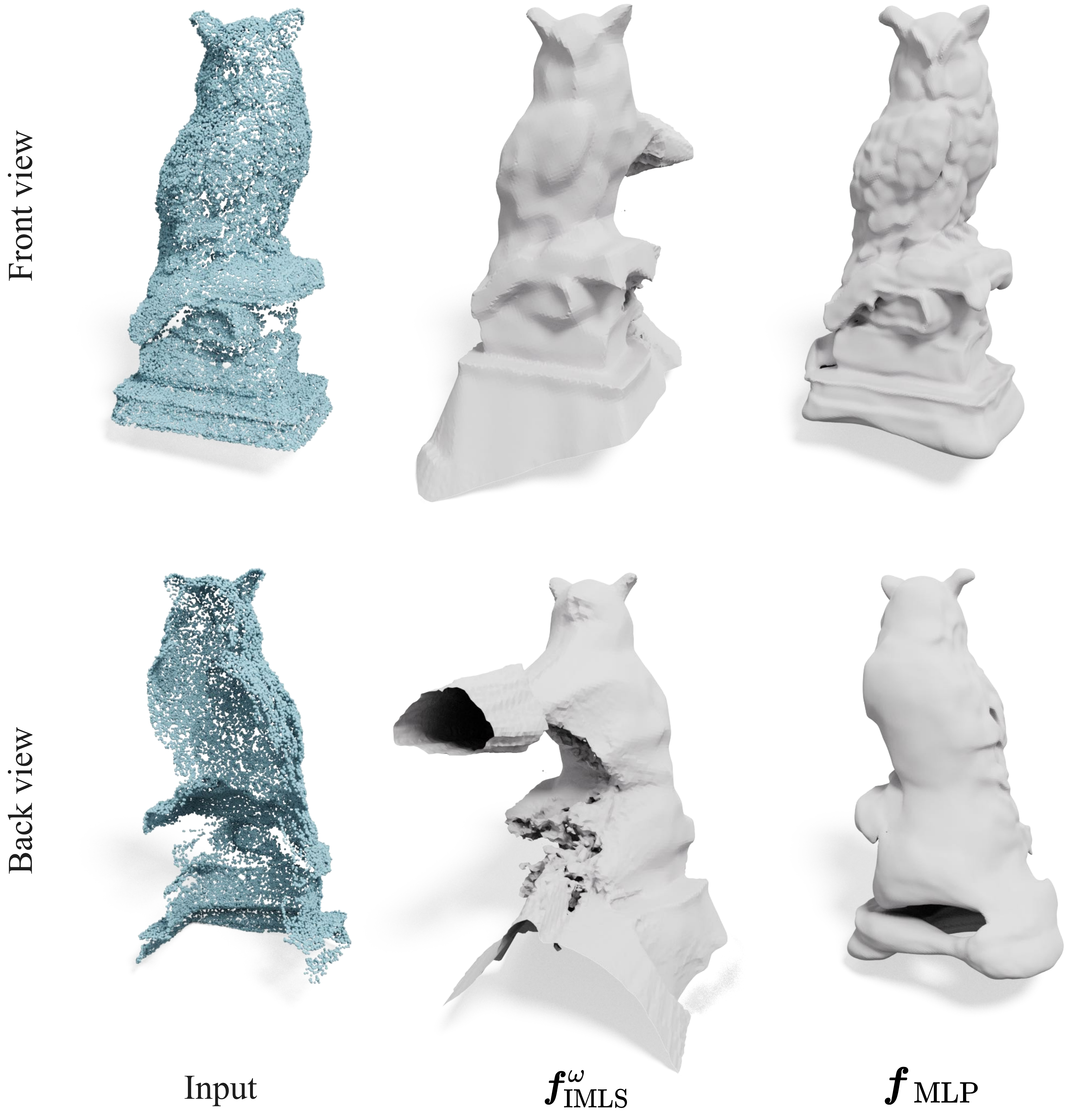}
    \vspace{-3mm}
    \caption{
    Comparing $\boldsymbol{f}^{\omega}_\text{IMLS}$ and $\boldsymbol{f}_\text{MLP}$ for scan with noise and missing parts.
    }
    \label{fig:abl_imls_g_f}
\end{figure}

\vspace{.5mm}
\noindent
\textbf{Comparison to existing IMLS methods.}
Our neural network inherits the noise-resistance ability of the IMLS. 
A natural question arises: 
is Neural-IMLS just a variant of the original IMLS and cannot defeat the original IMLS
in the noise-free situation?
To answer the question, we compare the following 5 implicit functions:
(1)~the implicit surface obtained by IMLS~\cite{kolluri2008imls},
(2)~the implicit one by RIMLS~\cite{oztireli2009RIMLS},
(3)~the implicit one by SPR~\cite{kazhdan2013screened},
(4)~the implicit one reported by IMLS when our neural network gets converged, i.e., $\boldsymbol{f}_\text{IMLS}^{\omega}$,
and
(5)~the implicit one reported by MLP when our neural network gets converged, i.e., $\boldsymbol{f}_\text{MLP}$.
We provide the ground truth normal for (1), (2), and (3).
The qualitative results and the quantitative results are shown in Fig.~\ref{fig:abl_imls}and Table.~\ref{tab:abl_imls}, respectively.
In the top row of Fig.~\ref{fig:abl_imls}, 
we show a teapot model whose point cloud is free of noise and complete.
As can be observed, the RIMLS easily produces broken parts, while the IMLS struggles to preserve local geometric features~(see the handle of the Utah teapot),
and RIMLS easily produces broken parts. 
In the bottom row of Fig.~\ref{fig:abl_imls}, 
we show a cup model whose point cloud is free of noise but has missing parts 
in the inner wall (around the handle). 
It can be seen that IMLS, RIMLS, and SPR suffer from the data-missing defect. 
Additionally, from Fig.~,\ref{fig:abl_imls_g_f} we can see that $\boldsymbol{f}_\text{MLP}$ has a stronger completion capability than  $\boldsymbol{f}_\text{IMLS}^{\omega}$ since the neural network, whose basis functions are globally defined, naturally owns the ability of shape completion.

\begin{table}[!htp]
\centering
\caption{\textbf{Comparison between different IMLS methods.}}
\label{tab:abl_imls}
\begin{tabular}{l|c} 
\toprule
           & FAMOUS no-n.   \\ 
\midrule
IMLS~\cite{kolluri2008imls}       & 2.92          \\
RIMLS~\cite{oztireli2009RIMLS}      & 1.46  \\
SPR~\cite{kazhdan2013screened}        & 1.53           \\
Func $\boldsymbol{f}^\omega_\text{IMLS}$  & 1.36           \\
MLP $\boldsymbol{f}_\text{MLP}$ & \textbf{1.34}  \\
\bottomrule
\end{tabular}
\end{table}

\vspace{.5mm}
\noindent
\textbf{Comparison to DeepMLS.}
DeepMLS~\cite{Liu2021MLS} is also a learning-based reconstruction method based on IMLS~\cite{kolluri2008imls}.
Like our approach, DeepMLS can also deal with noisy data,
but it needs to be trained with ground-truth SDFs. 
Next, we will compare our approach with  DeepMLS.

For a fair comparison, we reproduce the results of DeepMLS using the officially-released pre-trained model.
The test dataset is the low-noise FAMOUS, where the constant Gaussian noise strength
is set to 0.005 of the maximum side length of the bounding box,
which is consistent with the training data of the original DeepMLS. 

\begin{figure}[!htp]
    \centering
    \includegraphics[
      width=0.48\textwidth,
    ]{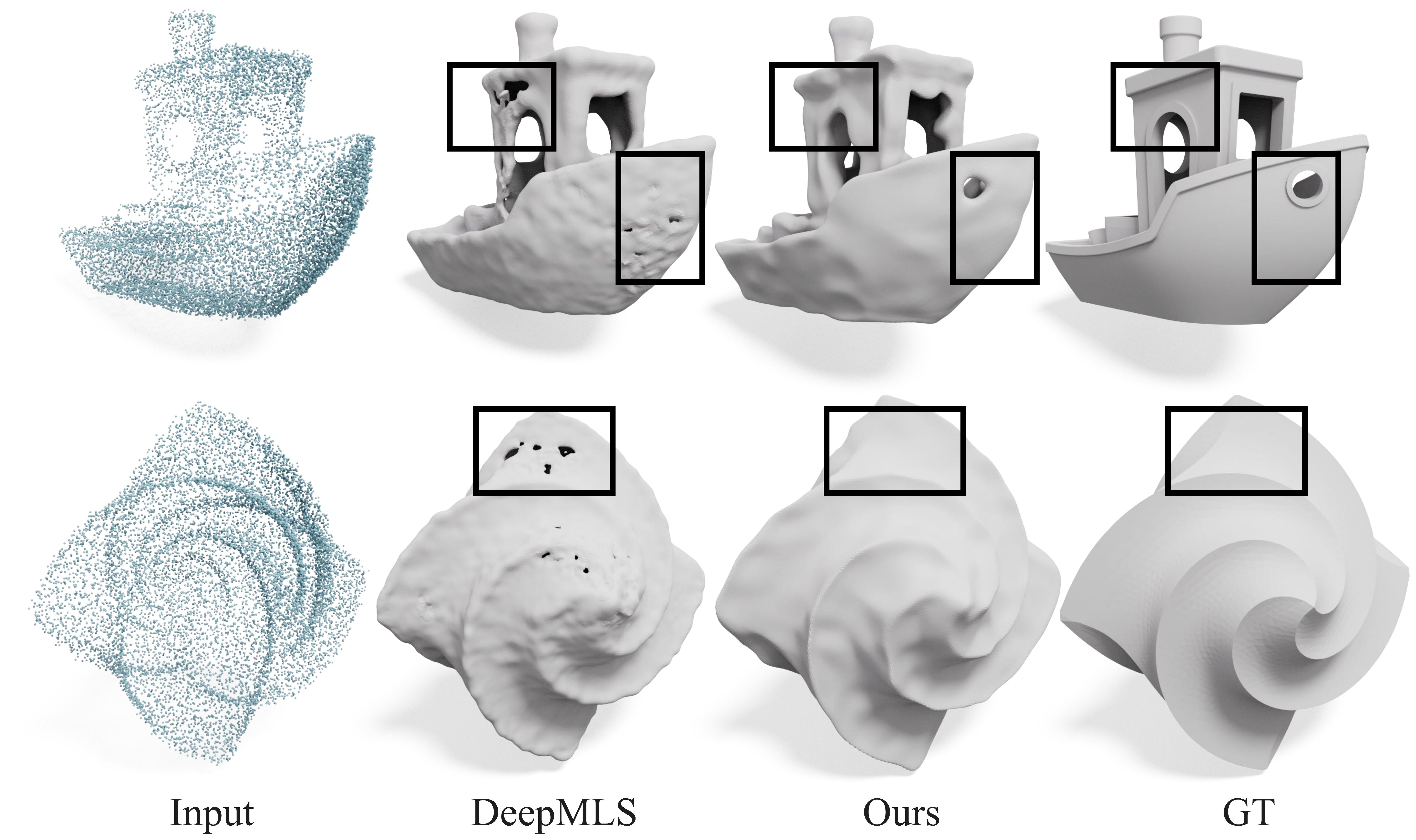}
    \vspace{-3mm}
    \caption{
    Comparison between DeepMLS~\cite{Liu2021MLS} and ours.
    }
    \label{fig:abl_deepmls}
\end{figure}

Tab.~\ref{tab:deep_comp} and Fig.~\ref{fig:abl_deepmls}
respectively present the quantitative results and the qualitative results. 
The disadvantages of DeepMLS are two-fold. 
First, DeepMLS cannot predict a suitable radius.
Second, DeepMLS does not take the spatial coherence of gradients into account. 


\begin{table}[!htp]
\centering
\caption{\textbf{Comparison between DeepMLS~\cite{Liu2021MLS} and ours.}}
\label{tab:deep_comp}
\begin{tabular}{l|c} 
\toprule
        & FAMOUS low-n  \\ 
\midrule
DeepMLS & 1.68           \\
\textbf{Ours}    & \textbf{1.54}  \\
\bottomrule
\end{tabular}
\end{table}

\vspace{.5mm}
\noindent
\textbf{Impact of Local Patch Size.}
We also conduct an ablation study about how the local patch size influences the
reconstruction results. 
Our tests are respectively made on the no-noise, medium-noise
and max-noise FAMOUS datasets.
The statistics are available in Table~\ref{tab:abl_localsize}.

Generally speaking, one has to increase the radius~$r$ 
or the number of the nearest neighboring points
so as to adapt to noise.
For this purpose, we set $r_{\text{small}} = 0.01d_{{P}}$, $r_{\text{medium}} = 0.03d_{{P}}$, and $r_{\text{large}} = 0.1d_{{P}}$, 
respectively on the three datasets. 
We also test the k-nearest neighbor algorithm (k-NN) with different $k$'s: $k_{\text{small}}=100$, $k_{\text{medium}}=300$, and $k_{\text{large}}=500$, respectively.

The experimental observations are two-fold.
On the one hand, controlling the radius is better than controlling the number of nearest neighbors since the former can deal with irregular point distributions. 
On the other hand, a larger $r$ tends to report a smoother surface.
\begin{table}[!htp]
\centering
\caption{\textbf{Ablation study of local size.} We show the Chamfer Distance
scores for different local-size parameters.}
\label{tab:abl_localsize}
\begin{tabular}{l|ccc} 
\toprule
                  & \multicolumn{3}{c}{FAMOUS}                     \\
                  & no-n.         & med-n.        & max-n.         \\ 
\midrule
$r_\text{small}$  & \textbf{1.35} & 2.67          & 10.63          \\
$r_\text{medium}$ & 1.92          & \textbf{1.62} & 6.75           \\
$r_\text{large}$  & 2.04          & 2.95          & \textbf{4.71}  \\
$k_\text{small}$  & 1.63          & 1.92          & 9.38           \\
$k_\text{medium}$ & 1.76          & 2.35          & 6.99           \\
$k_\text{large}$  & 2.01          & 2.81          & 4.82           \\
\bottomrule
\end{tabular}
\end{table}

\section{Conclusions and Limitations}
In this paper, we introduced Neural-IMLS, a method that learns a noise-resistant SDF, in a self-supervised fashion, directly from a raw point cloud without orienting normals.
Our method operates in a mutual-regularization mechanism,
which uses the MLP to provide normals to the IMLS while using the IMLS to provide distances to the MLP as the reference. 
On the one hand, Neural-IMLS inherits the noise-resistance ability of the IMLS,
outperforming the existing overfitting methods in the presence of noise.
On the other hand, Neural-IMLS, benefiting from the global representation ability of the 
neural network, is able to deal with various kinds of defects, such as thin gaps and missing parts.
Extensive experiments and ablation studies validate its effectiveness.

Our neural network, in its current form, 
includes a hyperparameter~$r$.
We have to manually tune~$r$ to adapt to different levels of noise.
In the future, we shall explore the possibility
of automatically learning the adaptive parameter~$r$ to achieve a good balance between 
the noise-resistance ability
and 
the geometric feature preserving ability.


%



\ifCLASSOPTIONcaptionsoff
  \newpage
\fi



%
\bibliographystyle{IEEEtran}
\bibliography{egbib}

\appendices
~\section{Experiment Details}
~\subsection{Network architectures}
\label{app:network}
We use the same network architecture as IGR~\cite{IGR}:
\begin{figure}[!htp]
    \centering
    \includegraphics[
      width=0.48\textwidth,
    ]{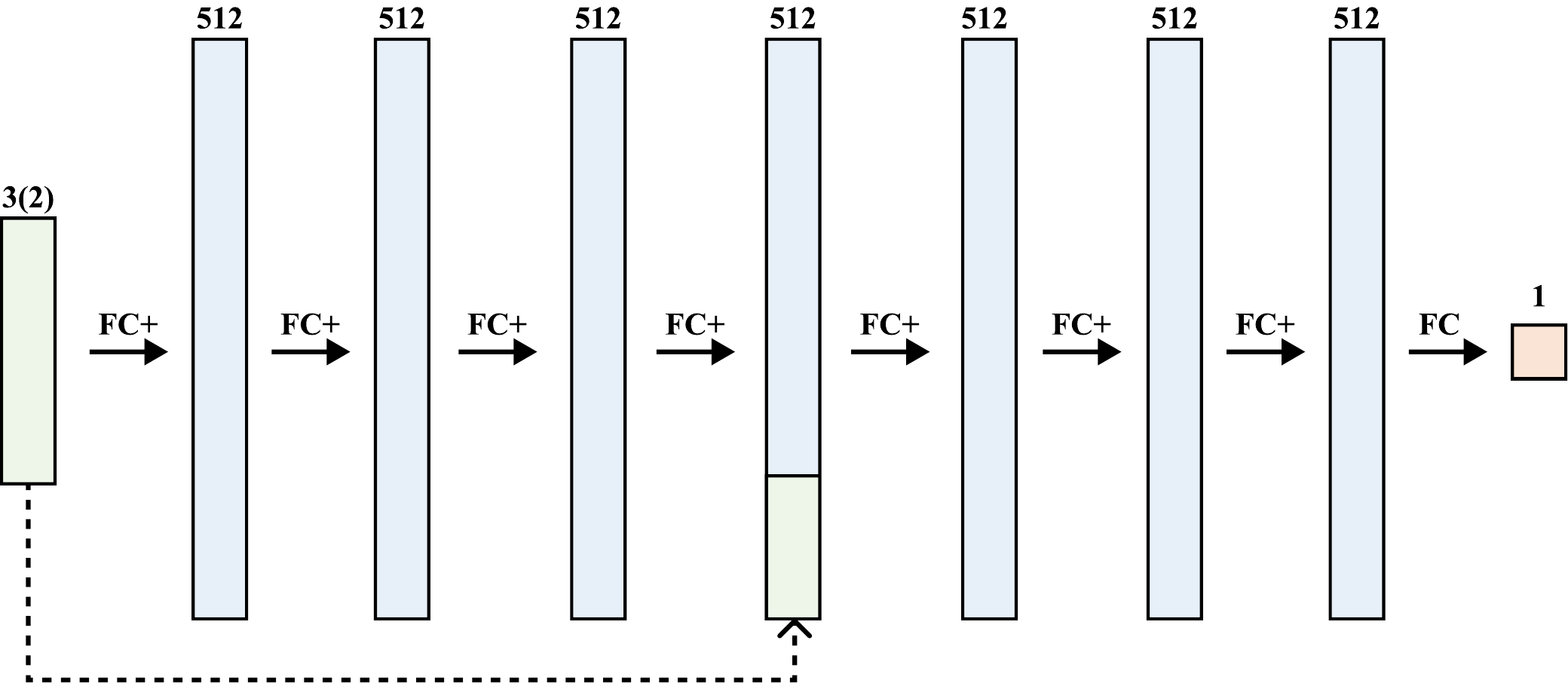}
    \caption{
    Network architectures.
    }
    \label{fig:network}
\end{figure}

\noindent
where input is 2D or 3D points and \textbf{FC} is a fully connected linear layer and \textbf{FC+} is \textbf{FC} followed by \textbf{Softplus} activation; a smooth approximation 1 of \textbf{ReLU}: $x \mapsto \frac{1}{\beta} \ln \left(1+e^{\beta x}\right)$. We use $\beta = 1000$ by default.
The dashed line connecting the input to the 4th layer indicates a skip connection. 

~\subsection{Parameters Settings}
We implement our method with PyTorch~\cite{paszke2019pytorch} under RTX3090.
We use Adam~\cite{adam} as the optimizer with a learning rate of~$5\text{e}^{-5}$ 
for training the network.
We group the query points in batches, of a batch size~100,
and set the max epoch value of our trainer to~200.
We sample 25 query points around each point~$\boldsymbol{p}_i$, following the Gaussian distribution,
which is the same sampling strategy as Neural-Pull~\cite{NeuralPull}. 
We align the standard deviation of the Gaussian distribution
with the distance from~$\boldsymbol{p}_i$ to the $50$-th nearest neighbor.

~\section{The experiments details}
\begin{table*}[!htp]
\centering
\caption{\textbf{Surface Reconstruction Benchmark.} 
We report Chamfer Distance (CD) Hausdorff distance (HD). All methods do not need ground-truth supervision and we evaluate them without normals. The \textit{scans} column reports the one-sided distances ($\vec{\text{CD}}$ and $\vec{\text{HD}}$) between the reconstruction and the simulated scans, which give a measure of the reconstruction's overfit to the noisy input.}
\label{table:add_srb}
\resizebox{\linewidth}{!}{%
\begin{tabular}{l|cc|cccc|cccc|cccc|cccc|cccc} 
\toprule
              & \multicolumn{2}{c|}{Mean}     & \multicolumn{4}{c|}{Anchor}                                & \multicolumn{4}{c|}{Daratech}                              & \multicolumn{4}{c|}{DC}                                    & \multicolumn{4}{c|}{Gargoyle}                              & \multicolumn{4}{c}{Lord Quas}                              \\
Method        & \multicolumn{2}{c|}{GT}       & \multicolumn{2}{c}{GT}        & \multicolumn{2}{c|}{Scans} & \multicolumn{2}{c}{GT}        & \multicolumn{2}{c|}{Scans} & \multicolumn{2}{c}{GT}        & \multicolumn{2}{c|}{Scans} & \multicolumn{2}{c}{GT}        & \multicolumn{2}{c|}{Scans} & \multicolumn{2}{c}{GT}        & \multicolumn{2}{c}{Scans}  \\
               & CD         & HD         & CD         & HD         & $\vec{\text{CD}}$ & $\vec{\text{HD}}$ & CD         & HD         & $\vec{\text{CD}}$ & $\vec{\text{HD}}$ & CD         & HD         & $\vec{\text{CD}}$ & $\vec{\text{HD}}$ & CD         & HD         & $\vec{\text{CD}}$ & $\vec{\text{HD}}$ & CD         & HD         & $\vec{\text{CD}}$ & $\vec{\text{HD}}$  \\ 
\midrule
SAL~\cite{SAL}          & 0.36          & 7.47          & 0.42          & 7.21          & 0.17        & 4.67         & 0.62          & 13.21         & 0.11        & 2.15         & 0.18          & 3.06          & 0.08        & 2.82         & 0.45          & 9.74          & 0.21        & 3.84         & 0.13          & 4.14          & 0.07        & 4.04         \\
IGR~\cite{IGR}          & 1.38          & 16.33         & 0.45          & 7.45          & 0.17        & 4.55         & 4.90          & 42.15         & 0.70        & 3.68         & 0.63          & 10.35         & 0.14        & 3.44         & 0.77          & 17.46         & 0.18        & 2.04         & 0.16          & 4.22          & 0.08        & 1.14         \\
IGR+FF~\cite{IGR}       & 0.96          & 11.06         & 0.72          & 9.48          & 0.24        & 8.89         & 2.48          & 19.6          & 0.74        & 4.23         & 0.86          & 10.3          & 0.28        & 3.98         & 0.26          & 5.24          & 0.018       & 2.93         & 0.49          & 10.7          & 0.14        & 3.71         \\
SIREN~\cite{sitzmann2020siren}        & 0.42          & 7.67          & 0.72          & 10.98         & 0.11        & 1.27         & 0.21          & 4.37          & 0.09        & 1.78         & 0.34          & 6.27          & 0.06        & 2.71         & 0.46          & 7.76          & 0.08        & 0.68         & 0.35          & 8.96          & 0.06        & 0.65         \\
PHASE+FF~\cite{lipman2021phase}     & 0.22          & 4.96          & 0.29          & 7.43          & 0.09        & 1.49         & 0.35          & 7.24          & 0.08        & 1.21         & 0.19          & 4.65          & 0.05        & 2.78         & \textbf{0.17} & 4.79          & 0.07        & 1.58         & \textbf{0.11} & \textbf{0.71} & 0.05        & 0.74         \\
SAP~\cite{peng2021shape}          & 0.20          & 4.60          & 0.34          & 8.25          & 0.10        & 3.88         & 0.22          & \textbf{3.00} & 0.07        & 1.76         & 0.17          & 3.22          & 0.06        & 0.66         & 0.18          & 5.21          & 0.07        & 0.68         & 0.13          & 3.42          & 0.05        & 0.92         \\
iPSR~\cite{hou2022iterative}         & 0.21          & 5.01          & 0.35          & 8.57          & 0.10        & 3.23         & 0.24          & 5.71          & 0.08        & 2.00         & 0.16          & 2.76          & 0.06        & 2.73         & 0.18          & 5.12          & 0.08        & 2.18         & 0.13          & 2.92          & 0.05        & 1.48         \\
DiGS~\cite{ben2022digs}         & 0.19          & 3.52          & 0.29          & 7.19          & 0.11        & 1.17         & \textbf{0.20} & 3.72          & 0.09        & 1.80         & \textbf{0.15} & \textbf{1.70} & 0.07        & 2.75         & \textbf{0.17} & 4.10          & 0.09        & 0.92         & 0.12          & 0.91          & 0.06        & 0.70         \\
PCP~\cite{ma2022surface}          & 0.54          & 7.11          & 0.80          & 8.12          & 0.36        & 2.38         & 0.72          & 8.75          & 0.53        & 2.91         & 0.37          & 3.83          & 0.30        & 2.48         & 0.50          & 8.47          & 0.35        & 3.23         & 0.35          & 6.43          & 0.23        & 3.34         \\
\textbf{Ours} & \textbf{0.18} & \textbf{3.15} & \textbf{0.22} & \textbf{4.66} & 7.78        & 8.25         & 0.21          & 4.75          & 7.50        & 7.74         & 0.16          & 2.62          & 0.53        & 0.95         & 0.19          & \textbf{2.88} & 5.79        & 6.38         & 0.12          & 0.86          & 1.43        & 1.66         \\
\bottomrule
\end{tabular}
}
\end{table*}

\begin{figure*}[!htp]
    \centering
    \includegraphics[
      width=0.8\textwidth,
    ]{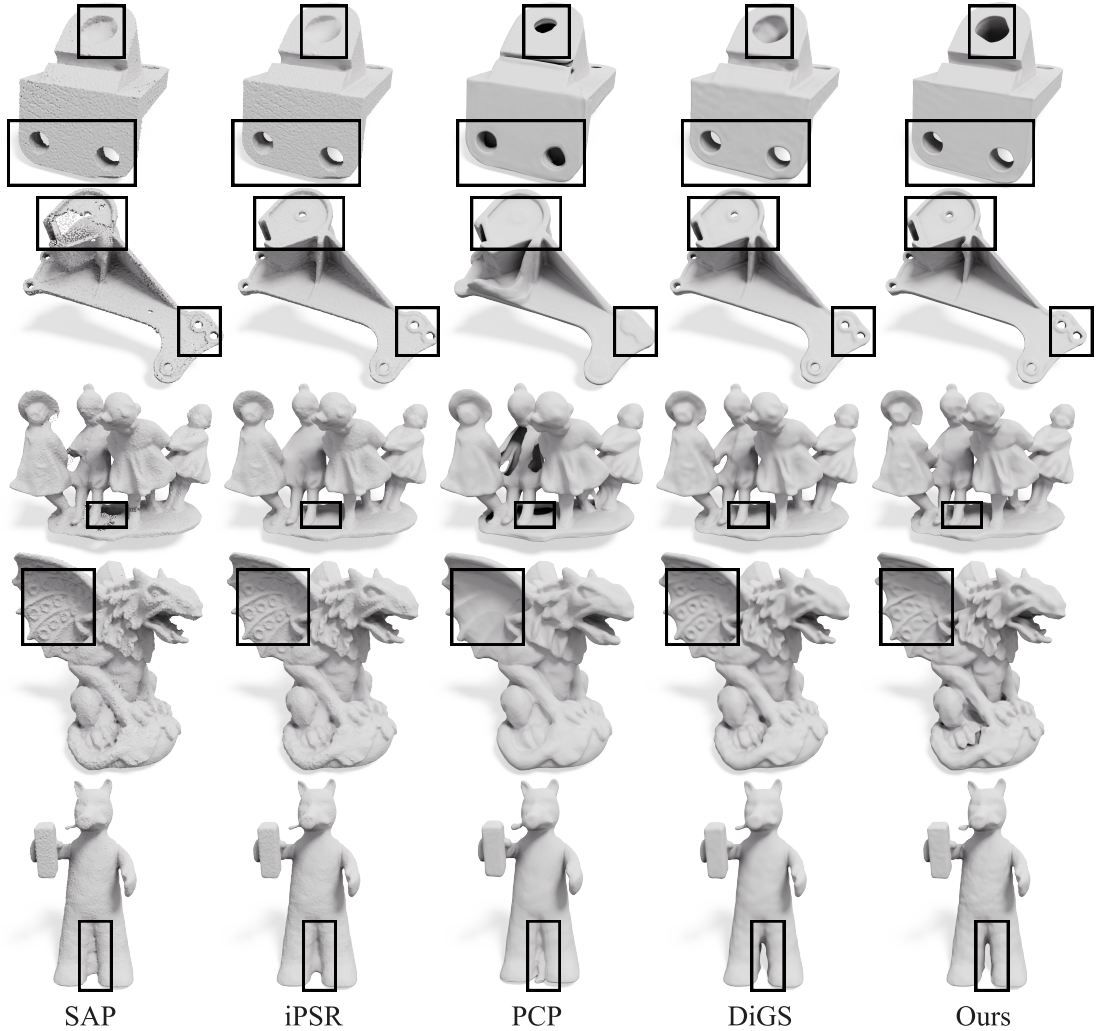}
    \vspace{-3mm}
    \caption{
    Qualitative comparison between different approaches on the Surface Reconstruction Benchmark~\cite{williams2019dgp}.}
    \label{fig:add_srb}
\end{figure*}

\begin{table*}[!htp]
\centering
\caption{\textbf{ShapeNet.} Quantitative results for surface reconstruction from
unoriented point clouds on the ShapeNet for each class.
RIMLS~\cite{oztireli2009RIMLS}, SPR~\cite{kazhdan2013screened}, and NSP~\cite{williams2021nsp} leverage oriented normals estimated by Dipole~\cite{metzer2021orienting}.}
\label{tab:add_shapenet}
\resizebox{\linewidth}{!}{%
\begin{tabular}{l|l|ccccccccccccc|c} 
\toprule
                                                            &               & airplane      & bench         & cabinet       & car           & chair         & display       & lamp          & loudspeaker   & rifle         & sofa          & table         & telephone     & watercraft    & Mean           \\ 
\midrule
\multirow{8}{*}{CD ($\times100$)$\downarrow$} & RIMLS~\cite{oztireli2009RIMLS}         & 0.64          & 1.01          & 1.30          & 0.91          & 1.26          & 1.14          & 1.54          & 1.35          & 0.37          & 1.23          & 1.42          & 0.58          & 0.86          & 1.04           \\
                                                            & SPR~\cite{kazhdan2013screened}           & 1.63          & 3.46          & 1.36          & 1.23          & 4.39          & 2.32          & 5.77          & 1.42          & 0.75          & 1.85          & 4.07          & 0.94          & 1.60          & 2.36           \\
                                                            & NSP~\cite{williams2021nsp}           & 3.07          & 1.87          & 1.29          & 0.73          & 2.69          & 1.31          & 2.45          & 1.15          & 0.57          & 1.21          & 2.92          & \textbf{0.43} & 1.22          & 1.60           \\
                                                            & SAL~\cite{SAL}           & 3.58          & 1.88          & 1.27          & 1.24          & 2.51          & 2.00          & 4.40          & 1.26          & 1.27          & 1.41          & 3.74          & 0.86          & 1.61          & 2.07           \\
                                                            & Neural-Pull~\cite{NeuralPull}   & 0.76          & 0.86          & 0.94          & 0.81          & 1.27          & 1.28          & 1.16          & \textbf{0.91} & 1.80          & 0.91          & 2.56          & 1.08          & 1.44          & 1.21           \\
                                                            & LPI~\cite{jiang2020local}           & 0.53          & 0.92          & 0.91          & 0.94          & 0.91          & 1.71          & 1.03          & 2.11          & 0.97          & 0.87          & 2.03          & 0.74          & 1.31          & 1.15           \\
                                                            & DiGS~\cite{ben2022digs}          & 0.57          & 0.80          & 1.57          & 0.82          & 1.19          & 1.11          & 0.99          & 1.42          & 0.56          & 1.21          & 1.48          & 0.80          & 0.82          & 1.02           \\
                                                            & \textbf{Ours} & \textbf{0.37} & \textbf{0.56} & \textbf{0.82} & \textbf{0.52} & \textbf{0.69} & \textbf{0.74} & \textbf{0.61} & 0.96          & \textbf{0.34} & \textbf{0.75} & \textbf{0.74} & 0.53          & \textbf{0.78} & \textbf{0.64}  \\ 
\midrule
\multirow{8}{*}{NC $\uparrow$}             & RIMLS~\cite{oztireli2009RIMLS} & 0.84          & 0.78          & 0.88          & 0.86          & 0.84          & 0.89          & 0.80           & 0.88          & 0.91          & 0.86          & 0.79          & 0.96          & 0.87          & 0.85           \\
                                                            & SPR~\cite{kazhdan2013screened}           & 0.78          & 0.72          & 0.87          & 0.83          & 0.73          & 0.84          & 0.74          & 0.88          & 0.85          & 0.84          & 0.71          & 0.92          & 0.84          & 0.81           \\
                                                            & NSP~\cite{williams2021nsp}           & 0.76          & 0.77          & 0.91          & 0.88          & 0.82          & 0.91          & 0.82          & 0.91          & 0.90          & 0.90          & 0.78          & \textbf{0.97} & 0.88          & 0.86           \\
                                                            & SAL~\cite{SAL}           & 0.73          & 0.77          & 0.92          & 0.87          & 0.82          & 0.89          & 0.75          & 0.92          & 0.83          & 0.90          & 0.80          & 0.96          & 0.85          & 0.84           \\
                                                            & Neural-Pull~\cite{NeuralPull}   & 0.84          & \textbf{0.92} & 0.92          & \textbf{0.89} & 0.89          & 0.93          & 0.85          & 0.93          & 0.89          & 0.92          & 0.85          & 0.95          & 0.86          & 0.89           \\
                                                            & LPI~\cite{jiang2020local}           & 0.87          & 0.86          & 0.92          & 0.88          & 0.91          & 0.91          & 0.87          & 0.90          & 0.90          & 0.92          & 0.86          & 0.95          & 0.87          & 0.89           \\
                                                            & DiGS~\cite{ben2022digs}          & 0.58          & 0.60          & 0.69          & 0.62          & 0.65          & 0.65          & 0.60          & 0.73          & 0.59          & 0.66          & 0.64          & 0.60          & 0.59          & 0.63           \\
                                                            & \textbf{Ours} & \textbf{0.92} & 0.87          & \textbf{0.94} & \textbf{0.89} & \textbf{0.92} & \textbf{0.95} & \textbf{0.88} & \textbf{0.94} & \textbf{0.93} & \textbf{0.93} & \textbf{0.91} & \textbf{0.97} & \textbf{0.90} & \textbf{0.91}  \\ 
\midrule
\multirow{8}{*}{FS $\uparrow$}             & RIMLS~\cite{oztireli2009RIMLS}         & 0.80          & 0.59          & 0.49          & 0.78          & 0.50          & 0.62          & 0.59          & 0.48          & 0.96          & 0.57          & 0.41          & 0.89          & 0.71          & 0.64           \\
                                                            & SPR~\cite{kazhdan2013screened}           & 0.55          & 0.48          & 0.55          & 0.77          & 0.33          & 0.51          & 0.38          & 0.58          & 0.88          & 0.61          & 0.29          & 0.88          & 0.70          & 0.57           \\
                                                            & NSP~\cite{williams2021nsp}           & 0.47          & 0.53          & 0.54          & 0.81          & 0.38          & 0.60          & 0.56          & 0.62          & 0.89          & 0.61          & 0.32          & \textbf{0.92} & 0.72          & 0.61           \\
                                                            & SAL~\cite{SAL}           & 0.31          & 0.51          & 0.57          & 0.70          & 0.46          & 0.49          & 0.39          & 0.53          & 0.60          & 0.6           & 0.34          & 0.76          & 0.58          & 0.52           \\
                                                            & Neural-Pull~\cite{NeuralPull}   & 0.71          & 0.61          & 0.63          & 0.69          & 0.61          & 0.62          & 0.50          & \textbf{0.69} & 0.47          & 0.61          & 0.46          & 0.72          & 0.53          & 0.60           \\
                                                            & LPI~\cite{LPI}          & 0.89          & 0.61          & 0.65          & 0.77          & 0.64          & 0.65          & 0.55          & 0.66          & 0.53          & 0.66          & 0.49          & 0.76          & 0.55          & 0.64           \\
                                                            & DiGS~\cite{ben2022digs}          & 0.83          & 0.73          & 0.41          & 0.70          & 0.57          & 0.60          & 0.62          & 0.37          & 0.80          & 0.53          & 0.46          & 0.71          & 0.68          & 0.61           \\
                                                            & \textbf{Ours} & \textbf{0.96} & \textbf{0.86} & \textbf{0.68} & \textbf{0.90} & \textbf{0.79} & \textbf{0.79} & \textbf{0.82} & 0.67          & \textbf{0.97} & \textbf{0.80} & \textbf{0.76} & 0.88          & \textbf{0.74} & \textbf{0.81}  \\
\bottomrule
\end{tabular}
}
\end{table*}
\begin{figure*}[!htp]
    \centering
    \includegraphics[
      width=\textwidth
    ]{shapenet_new.pdf}
    \vspace{-3mm}
    \caption{Qualitative comparison under ShapeNet~\cite{chang2015shapenet}. 
    All methods test on 10k noisy points and the oritented normals estimeted by Dipole~\cite{metzer2021orienting} for normal-based methods.}
    \label{fig:add_shapenet}
\end{figure*}
\subsection{Evaluation metrics}
\label{app:sub:evaluation}
To report the results on Surface Reconstruction Benchmark (SRB)~\cite{berger2013benchmark}, 
we use the Chamfer distance (CD) and Hausdorff distances (HD) from DGP~\cite{williams2019dgp}. 
We denote $M_g$ and $M_p$ as the ground-truth mesh (or ground-truth point clouds) and the mesh of the predicted result. 
$\mathcal{X}:=$ $\left\{\mathbf{x}_1, \ldots, \mathbf{x}_{N_g}\right\}$ and $\mathcal{Y}:=\left\{\mathbf{y}_1, \ldots, \mathbf{y}_{N_p}\right\}$ are randomly sample points on these two meshes (or point clouds), respectively. 
We deine $\mathcal{P}_{g 2 p}(\mathbf{x})=\arg \min _{\mathbf{y} \in \mathcal{Y}}\|\mathbf{x}-\mathbf{y}\|_2$ and $\mathcal{P}_{p 2 g}(\mathbf{y})=$ $\arg \min _{\mathbf{x} \in \mathcal{X}}\|\mathbf{x}-\mathbf{y}\|_2$. 
$N_g$ and $N_p$ are the numbers of sampled points of $M_g$ and $M_p$, respectively.
$\mathbf{n}(\cdot)$ denote an operator that returns he normal vector of a given point.

We define the Chamfer distance (CD) and one directional Chamfer distance ($\vec{\mathrm{CD}}$) as below:
\begin{equation}
\begin{aligned}
\mathrm{CD} (\mathcal{X}, \mathcal{Y}) =& \frac{1}{2 N_g} \sum_{i=1}^{N_g}\left\|\mathbf{x}_i-\mathcal{P}_{g 2 p}\left(\mathbf{x}_i\right)\right\|_2+\\
& \frac{1}{2 N_p} \sum_{i=1}^{N_p}\left\|\mathbf{y}_i-\mathcal{P}_{p 2 g}\left(\mathbf{y}_i\right)\right\|_2,\\
\vec{\mathrm{CD}} (\mathcal{X}, \mathcal{Y}) =& \frac{1}{N_g} \sum_{i=1}^{N_g}\left\|\mathbf{x}_i-\mathcal{P}_{g 2 p}\left(\mathbf{x}_i\right)\right\|_2.
\label{eq:cd}
\end{aligned}
\end{equation}
And Hausdorff distance (HD) and one directional Hausdorff distance ($\vec{\mathrm{HD}}$) as below:
\begin{equation}
\begin{aligned}
&\mathrm{HD}\left(\mathcal{X}, \mathcal{Y}\right)=\max \left(\vec{\mathrm{HD}}\left(\mathcal{X}, \mathcal{Y}\right),
\vec{\mathrm{HD}}\left(\mathcal{X}, \mathcal{Y}\right)\right),\\
&\vec{\mathrm{HD}}\left(\mathcal{X}, \mathcal{Y}\right)=\max_{\mathbf{x} \in \mathcal{X}} \min _{\mathbf{y} \in \mathcal{Y}}\left\|\mathbf{x}-\mathbf{y}\right\|_2
\end{aligned}
\end{equation}

For the ShapeNet dataset and Synthetic indoor dataset, we followed CONet~\cite{peng2020convolutional} not only report the Chamfer Distance as Eq.~\ref{eq:cd}, but also report the Normal Consistency (NC),
\begin{equation}
\begin{aligned}
\mathrm{NC}(\mathcal{X}, \mathcal{Y})=& \frac{1}{2 N_g} \sum_{i=1}^{N_g}\left|\mathbf{n}\left(\mathbf{x}_i\right) \cdot \mathbf{n}\left(\mathcal{P}_{g 2 p}\left(\mathbf{x}_i\right)\right)\right|+\\
& \frac{1}{2 N_p} \sum_{i=1}^{N_p}\left|\mathbf{n}\left(\mathbf{y}_i\right) \cdot \mathbf{n}\left(\mathcal{P}_{p 2 g}\left(\mathbf{y}_i\right)\right)\right|,
\end{aligned}
\label{eq:nc}
\end{equation}
and F-Score (FS), which is the harmonic mean between Precision and Recall. Precision is the percentage of points on $M_p$ that lie within distance $\tau$ to $M_g$, Recall is the percentage of points on $M_g$ that lie within distance $\tau$ to $M_p$:
\begin{equation}
\text{FS}(\mathcal{X}, \mathcal{Y})=\frac{2 * \text { precision } * \text { Recall }}{\text { precision }+\text { Recall }}.
\label{eq:fs}
\end{equation}

For ABC, FAMOUS and Thingi10k dataset, we use the Chamfer distance implementation of Points2Surf~\cite{erler2020points2surf}. 
We must point out that Point2Surf declares it use the squared Chamfer distance as below (Eq.10 in its main paper):
\begin{equation}
\begin{aligned}
\frac{1}{N_g} \sum_{i=1}^{N_g}\left\|\mathbf{x}_i-\mathcal{P}_{g 2 p}\left(\mathbf{x}_i\right)\right\|_2^2+
& \frac{1}{N_p} \sum_{i=1}^{N_p}\left\|\mathbf{y}_i-\mathcal{P}_{p 2 g}\left(\mathbf{y}_i\right)\right\|_2^2.
\label{eq:p2scd_paper}
\end{aligned}
\end{equation}

However, after correspondence with the authors of Points2Surf~\cite{p2scd}, we confirmed that they actually leverage below formulation in implementation followed:
\begin{equation}
\begin{aligned}
\sum_{i=1}^{N_g}\left\|\mathbf{x}_i-\mathcal{P}_{g 2 p}\left(\mathbf{x}_i\right)\right\|_2+
& \sum_{i=1}^{N_p}\left\|\mathbf{y}_i-\mathcal{P}_{p 2 g}\left(\mathbf{y}_i\right)\right\|_2,
\label{eq:p2scd_act}
\end{aligned}
\end{equation}
which is the varients of Eq.~\ref{eq:cd} without the mean operator to normalize by the number of sampled points.

For real-scanned benchmark, we followed~\cite{huang2022survey} to report Chamfer distance, Normal Consistency (NC), F-Score (FS) based Eq.~\ref{eq:cd},~\ref{eq:nc},~\ref{eq:fs}.
Moreover, we report Neural Feature Similarity (NFS) proposed by~\cite{huang2022survey}, focusing
on perceptual consistency between each reconstruction and the
ground-truth:

\begin{equation}
\mathrm{NFS} (\mathcal{X}, \mathcal{Y}) = \frac{\boldsymbol{h}_{\boldsymbol{\theta}}(\mathcal{X}) \cdot \boldsymbol{h}_{\boldsymbol{\theta}}(\mathcal{Y})}{\left\|\boldsymbol{h}_{\boldsymbol{\theta}}(\mathcal{X})\right\|_2\left\|\boldsymbol{h}_{\boldsymbol{\theta}}(\mathcal{Y})\right\|_2},
\end{equation}
where the feature-extracting network $\boldsymbol{h}_{\boldsymbol{\theta}}$ takes in a point cloud $\mathcal{X}$ and outputs a feature vector.


\subsection{Surface reconstruction on SRB}
\noindent
\textbf{Implementation details. }
We report the values for SAL~\cite{SAL}, IGR~\cite{IGR}, IGR+FF~\cite{tancik2020fourier}, SIREN~\cite{sitzmann2020siren}, PHASE+FF~\cite{ben2022digs}, DiGS~\cite{ben2022digs} from DiGS~\cite{ben2022digs}.
We report the results for SAP~\cite{peng2021shape}, iPSR~\cite{hou2022iterative}, PCP~\cite{ma2022surface} using their code and all method using $512^3$ grid same as ours.

\noindent
\textbf{Additional quantitative results.} We provide additional
quantitative results for surface reconstruction on the Surface
Reconstruction Benchmark~\cite{williams2019dgp}. 
Table~\ref{table:add_srb} shows errors for each shape.
Also, we report the one-sided distances to measure of the reconstruction’s overfit to the noisy input.
Our method does not cause overfitting in the presence of noise, which shows that our method has a noise-resistant ability.

\noindent
\textbf{Additional qualitative results.}
We provide additional qualitative results for surface reconstruction on the Surface Reconstruction Benchmark~\cite{williams2019dgp}. 
Fig.~\ref{fig:add_srb} shows visualizations of the output reconstruction of different methods. 

\begin{figure*}[!htp]
    \centering
    \includegraphics[
      width=.9\textwidth,
    ]{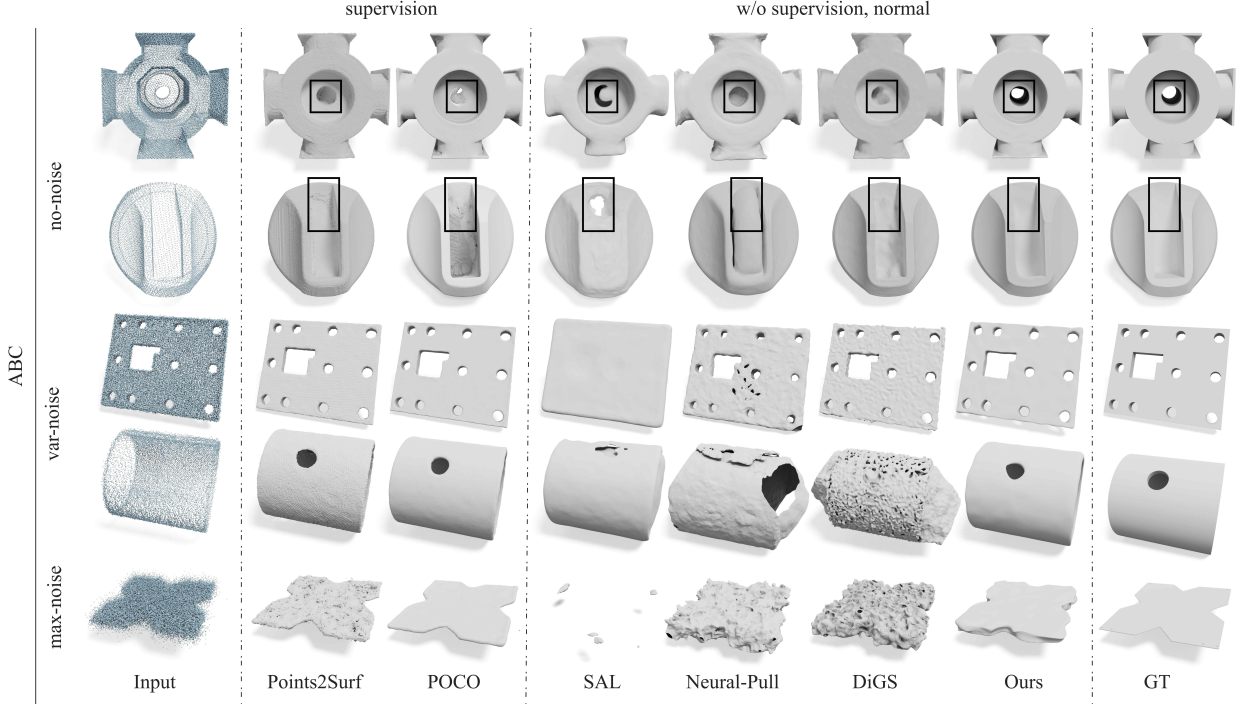}
    \vspace{-3mm}
    \caption{
    Qualitative comparison on the ABC dataset~\cite{koch2019abc},
    where the operations of
    sampling points from a mesh
    and adding noise to points follow~\cite{erler2020points2surf}. 
    }
    \label{fig:abc}
\end{figure*}

\subsection{Surface reconstruction on ShapeNet}
\noindent
\textbf{Implementation details.} We use the preprocessing, split and evaluation method from NSP~\cite{williams2021nsp}, and then sample 10k points for each shape with Gaussian noise that zero mean and 0.005 derivation.
We report the values for RIMLS~\cite{oztireli2009RIMLS}, SPR~\cite{kazhdan2013screened}
using the implentation of meshlab~\cite{cignoni2008meshlab}. 
The normals are estimated by Dipole~\cite{metzer2021orienting} with their pre-trained models.
We report the results for NSP~\cite{williams2021nsp}, SAL~\cite{SAL}, Neural-Pull~\cite{NeuralPull}, LPI~\cite{LPI}, and DiGS~\cite{ben2022digs} using their codes with $256^3$ grid same as ours.

\noindent
\textbf{Additional quantitative results.} We give the statistics of metrics per shape class in Table.~\ref{tab:add_shapenet}. 
As with the summary over all shape classes, we can see that we get better results among almost classes than other methods.

\noindent
\textbf{Additional qualitative results.} We show additional visualisations in Fig.~\ref{fig:add_shapenet}.
Our method get the better results even the noisy input with thin structures.


\subsection{Surface reconstruction on ABC, FAMOUS, Thingi10k}
\label{app:sub:abc}
\noindent
\textbf{Implementation details.} We report the values for DeepSDF~\cite{park2019deepsdf}, AatlasNet~\cite{groueix2018Aatlas}, SPR~\cite{kazhdan2013screened}, and Points2Surf~\cite{erler2020points2surf} from the main paper of Points2Surf~\cite{erler2020points2surf} and report values of POCO from their main paper~\cite{Boulch2022poco}.
We report the values for SAL~\cite{SAL}, Neural-Pull~\cite{NeuralPull}, and DiGS~\cite{ben2022digs} using their codes with $256^3$ grid.
\vspace{0.05mm}

\noindent
\textbf{Remark}: It worth noting that the results of Neural-Pull is inconsistently with its main paper.
After correspondence with the authors of Neural-Pull~\cite{NeuralPull}, we confirmed that they use the formulation of squared Chamfer distance formulation as below in implementation~\cite{pullcd}:
\begin{equation}
\begin{aligned}
&\frac{1}{2N_g} \sum_{i=1}^{N_g}\left\|\mathbf{x}_i-\mathcal{P}_{g 2 p}\left(\mathbf{x}_i\right)\right\|_2^2 + \\
& \frac{1}{2N_p} \sum_{i=1}^{N_p}\left\|\mathbf{y}_i-\mathcal{P}_{p 2 g}\left(\mathbf{y}_i\right)\right\|_2^2,
\end{aligned}
\end{equation}
which is different to Eq.~\ref{eq:p2scd_act} that Points2Surf actually used, leading to misleading in the main paper of Neural-Pull.
Specifically, the quantitative results reported by this formulation is much smaller than Eq.~\ref{eq:p2scd_act} thanks to the mean operator and squared operator.
Therefore, we use the evaluate codes of Points2Surf~\cite{erler2020points2surf} with no changes as POCO~\cite{Boulch2022poco} did for fairly comparison (We provide Neural-Pull~\cite{NeuralPull} evaluation results and ours under ABC no-n., ABC var-n., FAMOUS no-n., FAMOUS med-n., Thingi10K no-n., and Thingi10K med-n. in the supplemental material).

\begin{figure*}[!htp]
    \centering
    \includegraphics[
      width=0.9\textwidth,
    ]{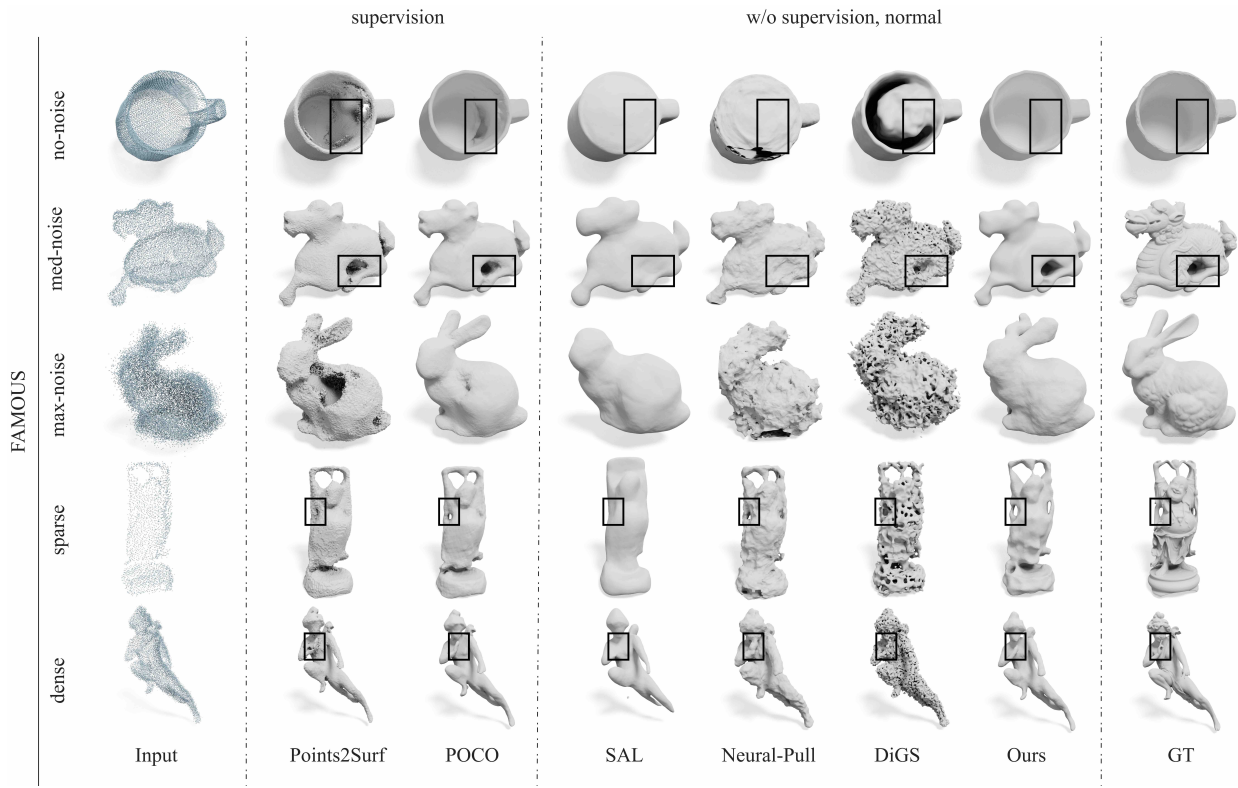}
    \vspace{-3mm}
    \caption{
    Qualitative comparison under FAMOUS.
    The input point clouds are generated following~\cite{erler2020points2surf}. 
    }
    \label{fig:famous}
\end{figure*}

\begin{figure*}[!htp]
    \centering
    \includegraphics[
      width=0.9\textwidth,
    ]{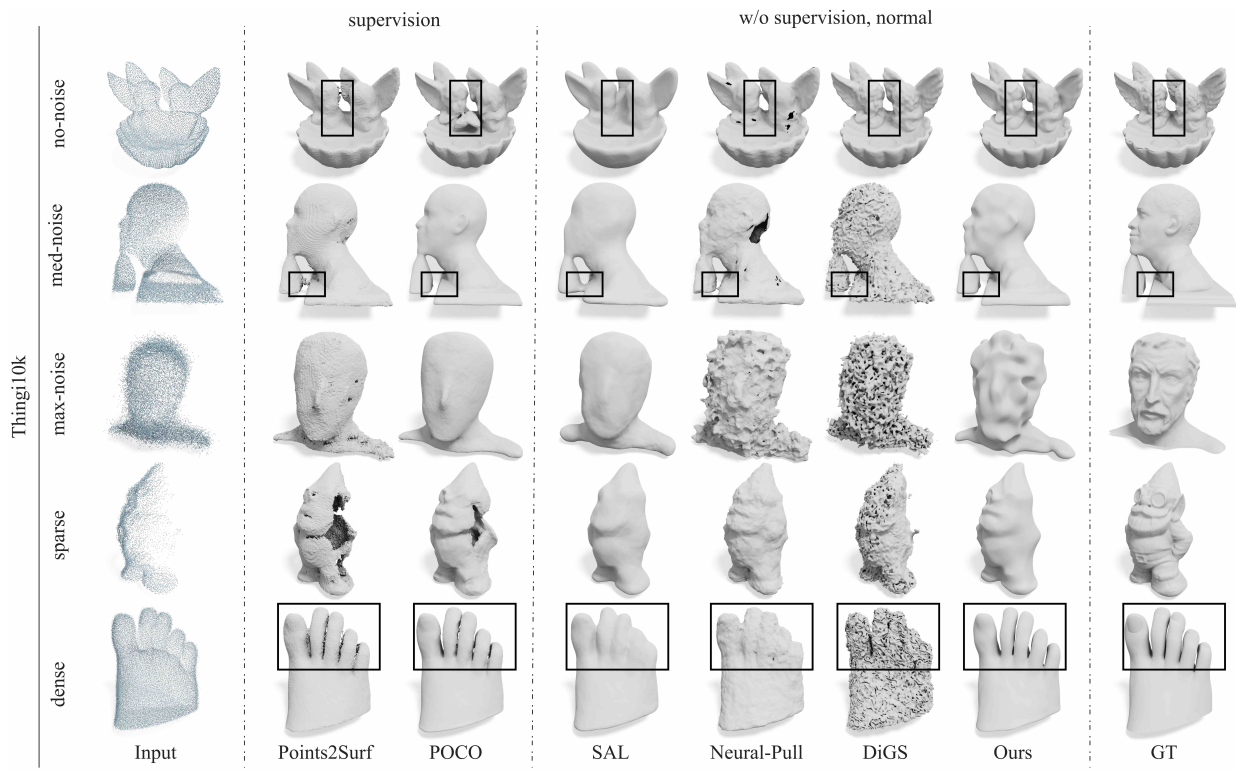}
    \vspace{-3mm}
    \caption{
    Qualitative comparison on the Thingi10K dataset~\cite{zhou2016thingi10k}.
     The input point clouds are generated following~\cite{erler2020points2surf}. 
    }
    \label{fig:thingi10k}
\end{figure*}

\noindent
\textbf{Qualitative results.} Fig.~\ref{fig:abc},~\ref{fig:famous},~\ref{fig:thingi10k} show qualitative results under ABC, FAMOUS, and Thingi10K datasets respectively.
Our method consistently show the better results than other overfitting methods.


\subsection{Surface reconstruction on Real-scanned benchmark}
\noindent
\textbf{Implementation details. } We report the values for BPA~\cite{bernardini1999ball}, SPR~\cite{kazhdan2013screened}, RIMLS~\cite{oztireli2009RIMLS}, IGR~\cite{IGR}, OccNet~\cite{mescheder2019onet}, DeepSDF~\cite{park2019deepsdf}, LIG~\cite{jiang2020local}, ParseNet~\cite{sharma2020parsenet}, Points2Surf~\cite{erler2020points2surf},
DSE~\cite{rakotosaona2021learning}, DeepMLS~\cite{Liu2021MLS}, GD~\cite{cohen2004greedy}, SALD~\cite{atzmon2021sald} from ~\cite{huang2022survey}.
And we report the values for PCP~\cite{ma2022surface}, DiGS~\cite{ben2022digs}, using their official codes.
We report POCO~\cite{Boulch2022poco} results with its pre-trained model under ABC dataset .

\subsection{Surface reconstruction on Synthetic indoor scene dataset}
We report the values for SAL~\cite{SAL}, IGR~\cite{IGR}, Neural-Pull~\cite{NeuralPull}, PCP~\cite{ma2022surface},  DiGS~\cite{ben2022digs} and SA-CONet~\cite{tang2021sa} using their codes with $512^3$ grid.

\section{Illustrative Examples}
\begin{figure}[!htp]
    \centering
    \includegraphics[
      width=0.48\textwidth,
    ]{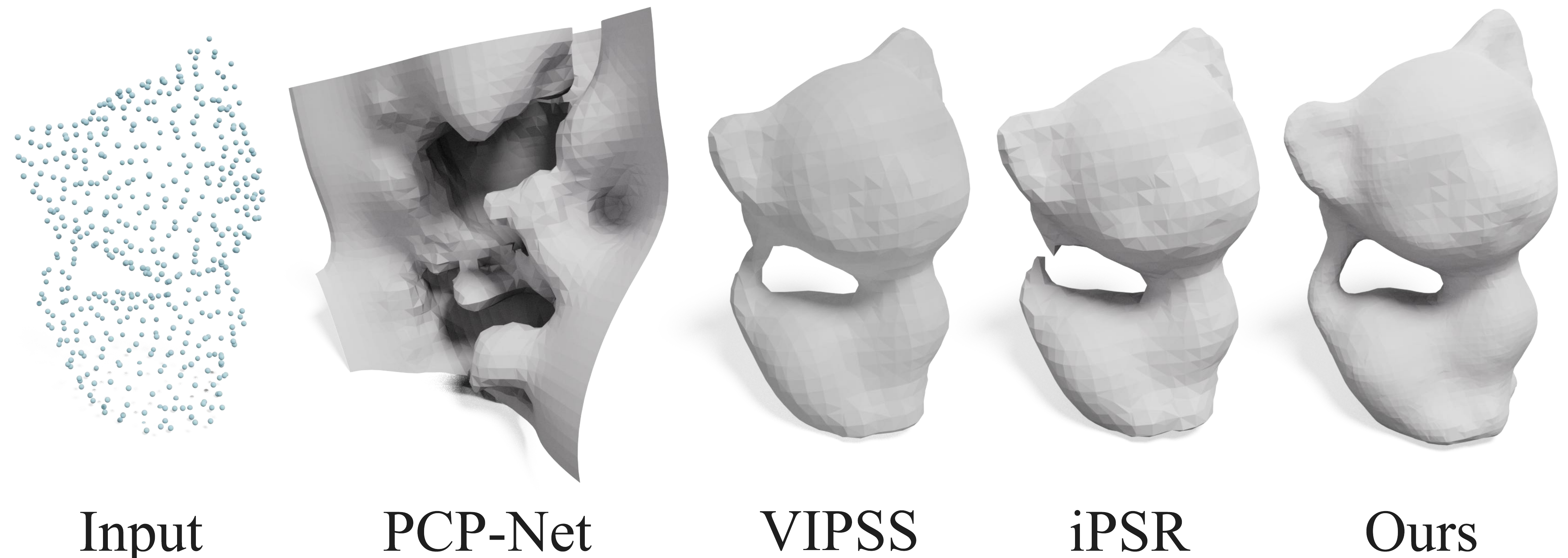}
    \vspace{-3mm}
    \caption{
    We use various approaches to generate normals,
    followed by feeding normals to SPR~\cite{kazhdan2013screened},
    to observe which kind of normals is more faithful.
    }
    \label{fig:normal}
\end{figure}

\noindent
\textbf{Oriented Normal Estimation.}
It is necessary to evaluate whether the normals produced by our approach are faithful.
For this purpose, we use various approaches to generate normals,
    followed by feeding normals to SPR~\cite{kazhdan2013screened},
    to evaluate the reconstruction quality. 
Specially, the approaches used to  generate normals
include PCPNet~\cite{PCP}, VIPSS~\cite{huang2019vipss}, iPSR~\cite{hou2022iterative}, and ours.
According to the viusal results shown in Fig.~\ref{fig:normal},
PCPNet fails due to the severe point insufficiency,
iPSR fails due to the incorrect inference about the normals around the thin tail,
and VIPSS is able to reconstruct the surface but the geometric features are not so prominent as ours. 
By contrast, ours is the best among the results on the super-sparse point cloud,
achieving a high fidelity around the tail and the ear of the kitten model,
which shows that our normals are more faithful.

%

\begin{figure}[!htp]
    \centering
    \includegraphics[
      width=0.48\textwidth,
    ]{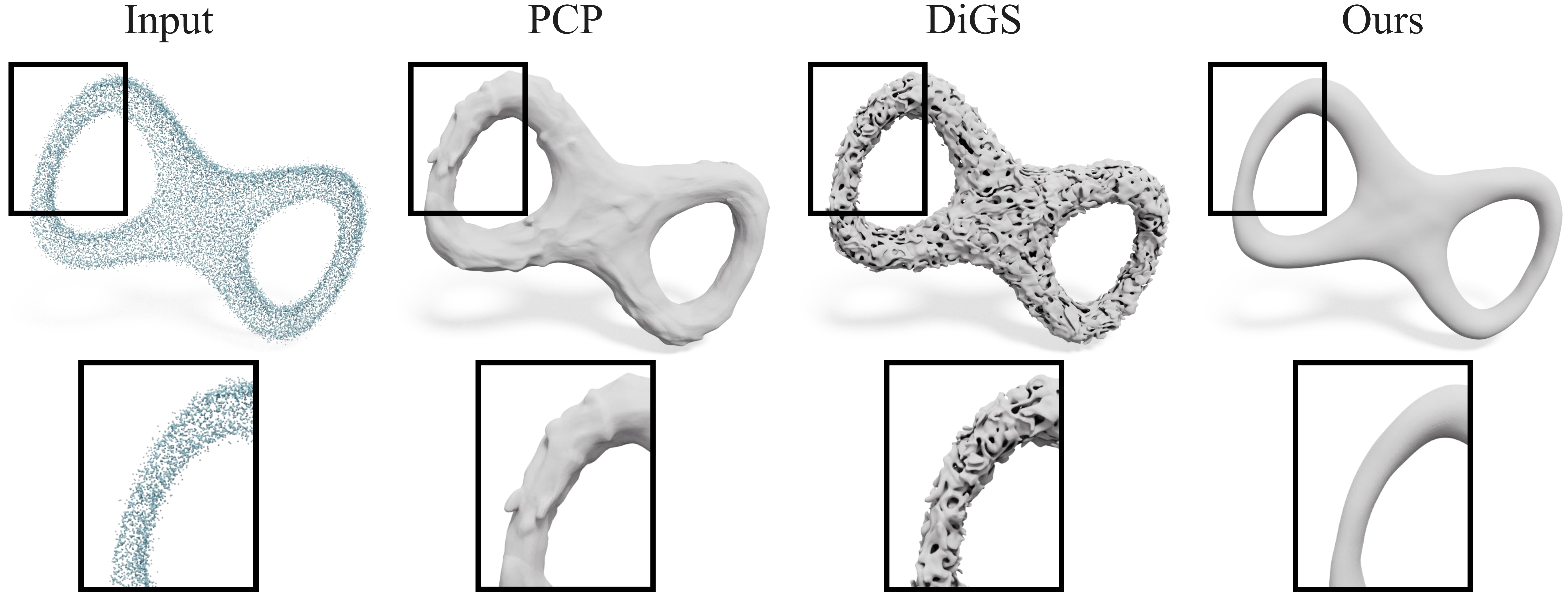}
    \vspace{-3mm}
    \caption{
    The reconstruction results from a noisy point cloud sampled from the implicit Bretzel. The existing overfitting methods, e.g., PCP~\cite{ma2022surface} and DiGS~\cite{ben2022digs}, do not resist noise.
    }
    \label{fig:implicit}
\end{figure}

\vspace{.5mm}
\noindent
\textbf{Geometric fidelity.}
To demonstrate the fidelity preserving ability of our
method, we conduct experiments on the noisy point cloud sampled from the implicit Bretzel, and compare our approach with PCP~\cite{ma2022surface} and DiGS~\cite{ben2022digs}.
The visual comparison (see Fig.~\ref{fig:implicit}) shows that existing SOTA overfitting  methods cannot produce a faithful result for a noisy input.
By contrast, our algorithm 
is superior to them
in terms of the ability to reconstruct
a high-fidelity surface even in the occurrence of noise.

\begin{figure}[!htp]
    \centering
    \includegraphics[
      width=0.48\textwidth,
    ]{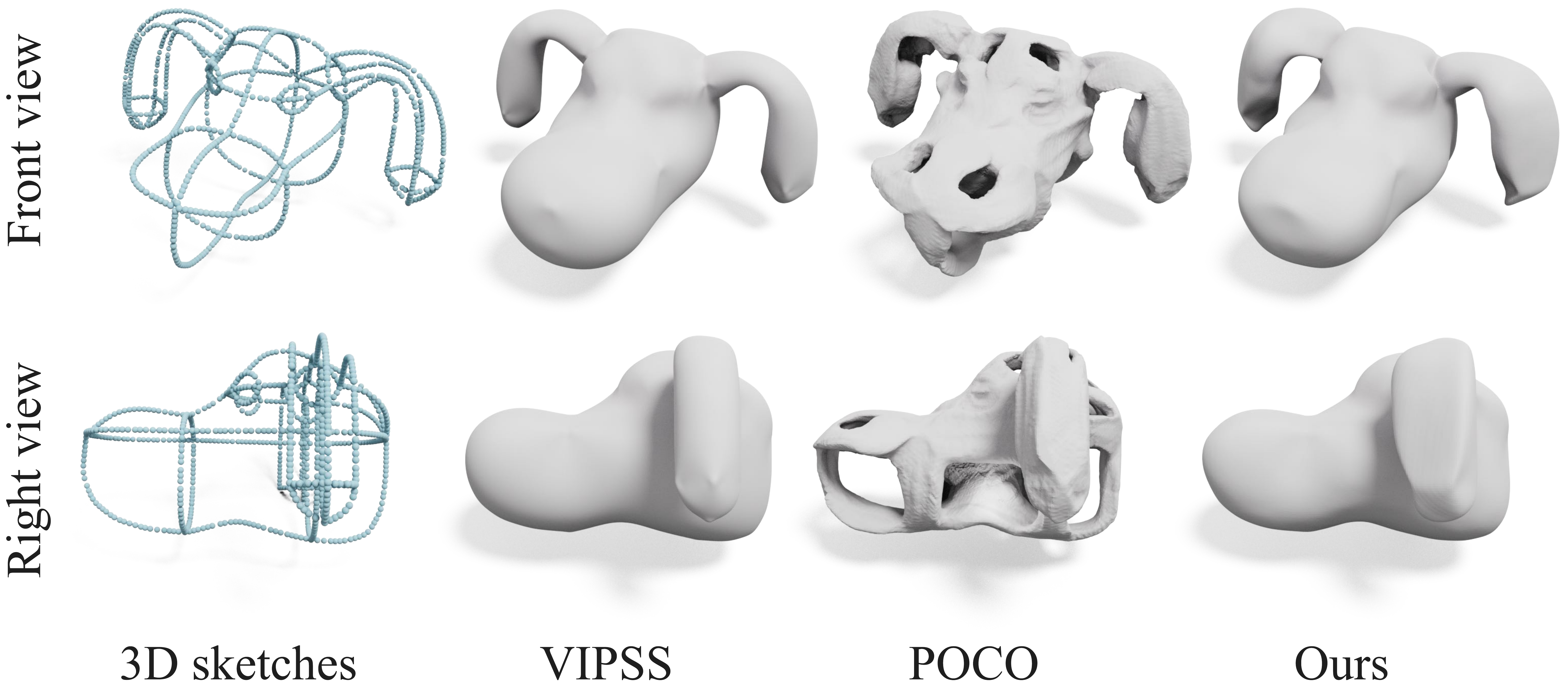}
    \vspace{-3mm}
    \caption{
    The reconstruction results from sketch points by different methods.
    }
    \label{fig:sketches}
\end{figure}

\vspace{.5mm}
\noindent
\textbf{3D Sketch.}
It is interesting 
to make clear if our approach can transform a super-sparse 3D sketch point cloud 
into a meaningful shape. 
In VIPSS~\cite{huang2019vipss},
the authors gave a 3D sketch point cloud of 1k points.
We visualize the reconstructed results by VIPSS~\cite{huang2019vipss},
POCO~\cite{Boulch2022poco} and ours
in Fig.~\ref{fig:sketches}.
It can be seen that POCO fails since their training set does not include the sketch-type data.
Our result is comparable to VIPSS
but VIPSS seriously suffers from the number of points. 

\begin{figure}[!htp]
    \centering
    \includegraphics[
      width=0.48\textwidth,
    ]{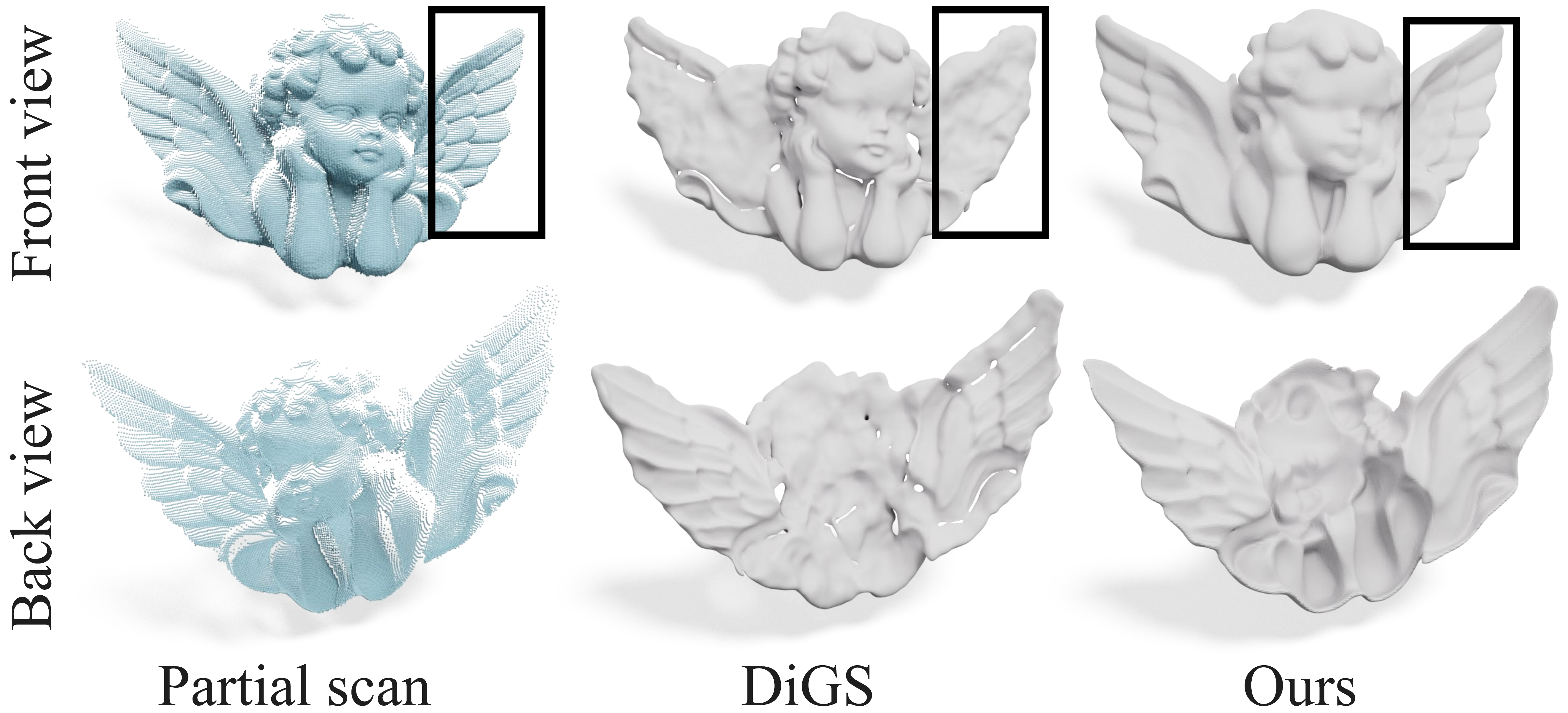}
    \vspace{-3mm}
    \caption{
    The reconstruction results on a partial-scan angel model. 
    Our method can not only recover the details of the shape but also fill the gaps. 
    Our neural network tends to reconstruct a closed shape even if the input point cloud has a single layer.}
    \label{fig:partial}
\end{figure}

\vspace{.5mm}
\noindent
\textbf{Partial Scan.}
By taking a partial scan as the input,
we come to observe the reconstruction ablity of our approach. 
In Fig.~\ref{fig:partial},
we use an angel model~\cite{calakli2011ssd}
to compare DiGS~\cite{ben2022digs} and ours. 
It can be seen that 
DiGS cannot fill the small gaps
but our approach outputs a closed model. 
Our explanation is that our neural network aims at reconstructing a closed shape even if the input point cloud has a single layer.


\begin{figure}[!htp]
    \centering
    \includegraphics[
      width=0.48\textwidth,
    ]{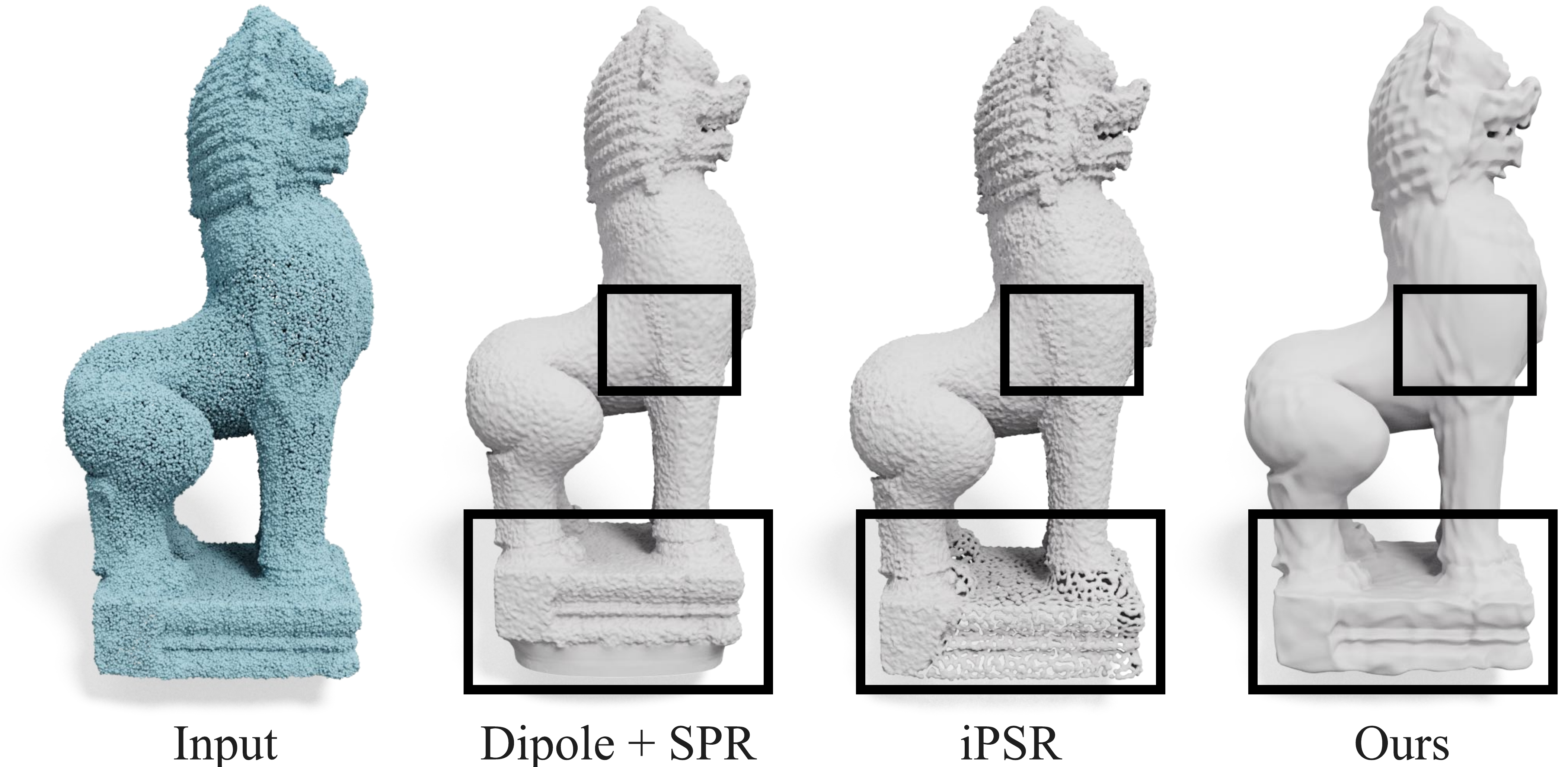}
    \vspace{-3mm}
    \caption{
    The reconstruction results on a large-size noisy point cloud (700k points). 
    Our method effectively resists noise, yielding a smooth surface.
    }
    \label{fig:large}
\end{figure}

\vspace{.5mm}
\noindent
\textbf{Large-size Point Cloud.} 
In order to test the ability of dealing with a large-size point cloud,
we use a real-scan~\cite{three_dscans} as the input, where the number of points amounts to 700k.
The approaches used for comparison include
(1)~SPR~\cite{kazhdan2013screened} plus Dipole~\cite{metzer2021orienting},
(2)~iPSR~\cite{hou2022iterative},
and (3)~ours. 
From Fig.~\ref{fig:large},
we can see that our approach can deal with a large-size noisy point cloud,
finally yielding a smooth surface.


\end{document}